\documentclass[10pt,twocolumn,letterpaper]{article}

\usepackage[pagenumbers]{cvpr} 

\usepackage{graphicx}
\usepackage{amsmath}
\usepackage{amssymb}
\usepackage{booktabs}
\usepackage{tabularx}
\usepackage{multicol}
\usepackage{multirow}
\usepackage{amsmath}
\usepackage{tablefootnote}

\usepackage[pagebackref,breaklinks,colorlinks]{hyperref}

\usepackage[capitalize]{cleveref}
\crefname{section}{Sec.}{Secs.}
\Crefname{section}{Section}{Sections}
\Crefname{table}{Table}{Tables}
\crefname{table}{Tab.}{Tabs.}

\def\cvprPaperID{10580} 
\def\confName{CVPR}
\def\confYear{2023}

\begin{document}

\title{Synthesizing Coherent Story with Auto-Regressive Latent Diffusion Models}
\author{
  Xichen Pan$^1$, Pengda Qin$^2$, Yuhong Li$^2$, Hui Xue$^2$, Wenhu Chen$^{1,3}$\\
  $^1$University of Waterloo, $^2$Alibaba Group, $^3$Vector Institute\\
}

\twocolumn[{
\maketitle
\vspace{-20pt}
\begin{center}
\captionsetup{type=figure}
\includegraphics[width=0.88\linewidth]{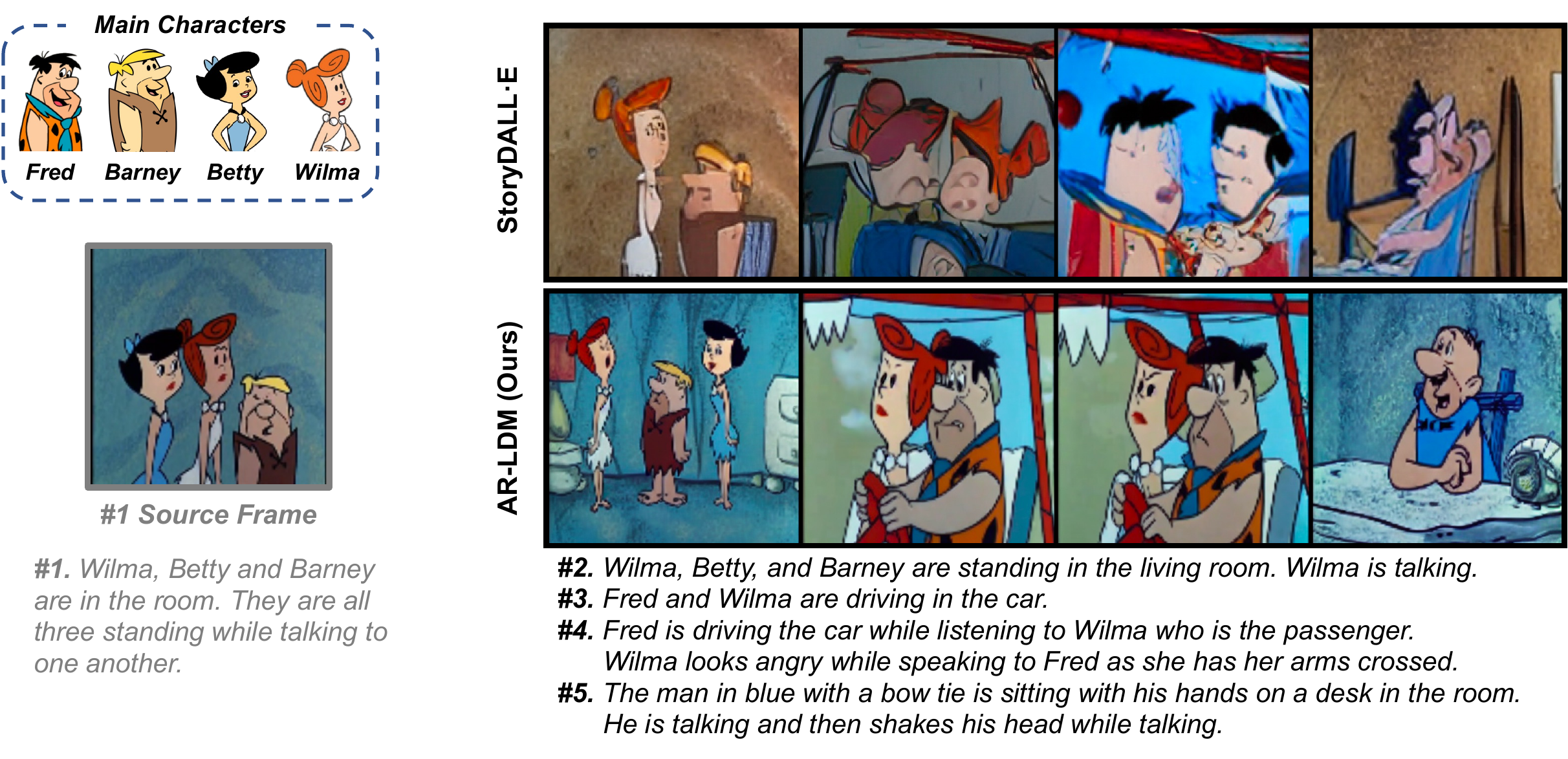}
\captionof{figure}{Comparison of a visual story example synthesized by AR-LDM (Ours) and StoryDALL·E~\cite{storydalle} on FlintstonesSV story continuation dataset. The visual stories are generated with reference to the source frame and captions.}
\label{fig:teaser}
\end{center}
}]


\renewcommand{\thefootnote}{\fnsymbol{footnote}}
\footnotetext{Technical Report.}
\renewcommand{\thefootnote}{\arabic{footnote}}

\begin{abstract}
Conditioned diffusion models have demonstrated state-of-the-art text-to-image synthesis capacity. Recently, most works focus on synthesizing independent images; While for real-world applications, it is common and necessary to generate a series of coherent images for story-stelling. In this work, we mainly focus on story visualization and continuation tasks and propose AR-LDM, a latent diffusion model auto-regressively conditioned on history captions and generated images. Moreover, AR-LDM can generalize to new characters through adaptation. To our best knowledge, this is the first work successfully leveraging diffusion models for coherent visual story synthesizing. Quantitative results show that AR-LDM achieves SoTA FID scores on PororoSV, FlintstonesSV, and the newly introduced challenging dataset VIST containing natural images. Large-scale human evaluations show that AR-LDM has superior performance in terms of quality, relevance, and consistency\footnote{Code available at \url{https://github.com/Flash-321/ARLDM}.}.
\end{abstract}

\section{Introduction}
\label{sec:introduction}
Recently advanced diffusion models~\cite{diffusionmodel} such as DALL·E 2~\cite{dalle2}, Imagen~\cite{imagen}, and Stable Diffusion~\cite{ldm} have shown unprecedented text-to-image synthetic capacities. These models focus on single-image generation, while many real-world use cases like comic drawing require models to generate a series of coherent images according to a long story description. Text-to-image models offer extreme freedom to guide creation through natural language. Simply generating each image according to every single caption will result in poor relevance and consistency. Textual Inversion~\cite{textualinversion} and DreamBooth~\cite{dreambooth} have focused on creating a specific unique concept to guide consistent generation results across images. Re-Imagen~\cite{reimagen} is able to generate specific entities with reference to retrieved image text pairs in a training-free manner. However, how to generate a series of coherent images illustrating a multi-sentence paragraph is still underexplored.

In this paper, we mainly focus on two tasks: story visualization~\cite{storygan} and story continuation~\cite{storydalle}. Story visualization aims at synthesizing a series of images to describe a story containing multiple sentences. Story continuation is a variant of story visualization with the same goal as story visualization, but additionally based on a source frame (i.e., the first frame). This setting addresses the generalization issue and limited information issue in story visualization, allowing models to generate more meaningful and coherent images. Story visualization and continuation are challenging tasks requiring both vision-language understanding and image generation. Previous works are mainly based on GANs and auto-regressive models, and utilize contextual text encoders to improve consistency. While, as the saying goes, ``\textit{A picture is worth a thousand words,}'' it is impossible for a single caption to exploit all necessary information for image generation. There are thousands of reasonable illustrations for a given story. For example, for the story shown in \cref{fig:teaser}, the captions of the third and fourth frames do not describe the detail of the ``\textit{car}'' or background. The key to generating coherent stories is to preserve as many details across images as possible. The main limitation of existing work is that the generation is guided only by contextual text conditions without leveraging previously generated images.

In this work, we propose Auto-Regressive Latent Diffusion Model (AR-LDM) to leverage diffusion models to synthesize coherent stories. Specifically, we employ a history-aware encoding module containing a CLIP text encoder~\cite{clip}, and a BLIP multimodal encoder~\cite{blip}. For each frame, AR-LDM is not only guided by the current caption but also conditioned on previously generated image-caption history. This allows AR-LDM to generate relevant and coherent images. As shown in \cref{fig:teaser}, AR-LDM shows strong multimodal understanding and image generation ability. It is able to precisely generate the scene as captions described in high quality, as well as keeping a strong consistency across frames. Additionally, we also explore adapting AR-LDM to preserve consistency for unseen characters (i.e., characters referred by a pronoun, like the man in the last frame of \cref{fig:teaser}) within the stories. This adaptation can largely alleviate the inconsistent generation results caused by uncertain descriptions of unseen characters.

To evaluate our method, we utilize two widely accepted datasets, FlintstonesSV and PororoSV, as our test bed. While all existing story visualization and continuation datasets are cartoon images\footnote{the DiDeMoSV~\cite{storydalle} dataset is a cartoon-style real-world dataset}, we introduce a new dataset VIST~\cite{vist} to better evaluate real-world story synthesis capacity. VIST contains story-in-sequence (SIS) captions that better match real-world use cases, and also provides description-in-isolation (DII) style captions. Quantitative evaluation results show our method achieves SoTA performance in both story visualization and continuation tasks. In particular, AR-LDM achieves an FID score of 16.59 on PororoSV, with a relative improvement of 70\% over previous story visualization methods. AR-LDM also boosts story continuation performance with a relative improvement of approximately 20\% on all evaluation datasets. We also conduct large-scale human evaluations to test our method's visual quality, relevance, and consistency, which shows that humans mostly prefer our synthesized stories over previous methods.

In general, our contribution can be summarized as follows:
\begin{enumerate}
    \item We propose a history-aware auto-regressive conditioned latent diffusion model AR-LDM, which first successfully leverages diffusion models for story synthesis.
    \item We introduce the VIST dataset and show AR-LDM is capable of real-world story synthesis.
    \item For more practical application, we additionally propose a simple but efficient adaptation method to allow AR-LDM generalizing to unseen characters.
\end{enumerate}
\section{Related work}
\subsection{Text-to-Image Synthesis}
Recent advances in text-to-image synthesis mainly focus on generative adversarial networks (GANs)~\cite{gan}, auto-regressive models, and diffusion models. GANs like Stackgan~\cite{stackgan}, Attngan~\cite{attngan}, Mirrorgan~\cite{mirrorgan}, and MXC-GAN~\cite{mxcgan} perform adversarial training between generators and discriminators to learn to generate high-quality images. Large auto-regressive models like DALL·E~\cite{dalle}, Make-A-Scene~\cite{makeascene}, and Parti~\cite{parti} can be easily scaled up and have also shown their excellent image synthetic capacity. Recently, success in diffusion models has attracted many researchers' attention. As likelihood-based models, diffusion models do not suffer from mode-collapse and potentially unstable training as GANs, and can generate more diversified images. Additionally, diffusion models are more parameter-effective than auto-regressive models.

\begin{figure*}[!th]
\begin{subfigure}{0.28\linewidth}
\centering
\includegraphics[height=155pt]{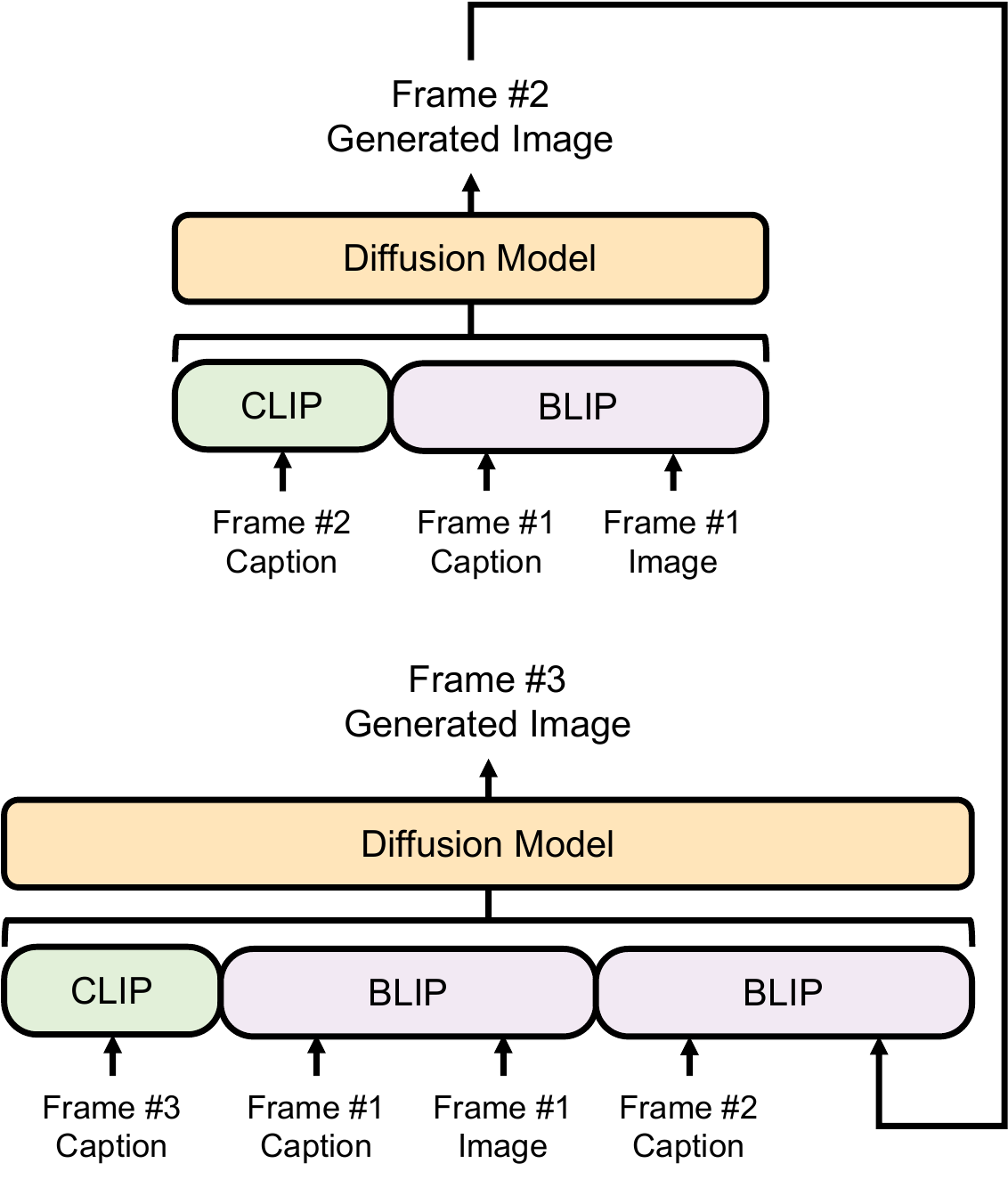}
\caption{Auto-regression process.}
\label{fig:ar}
\end{subfigure}
\hfill
\vline
\hfill
\begin{subfigure}{0.71\linewidth}
\centering
\includegraphics[height=155pt]{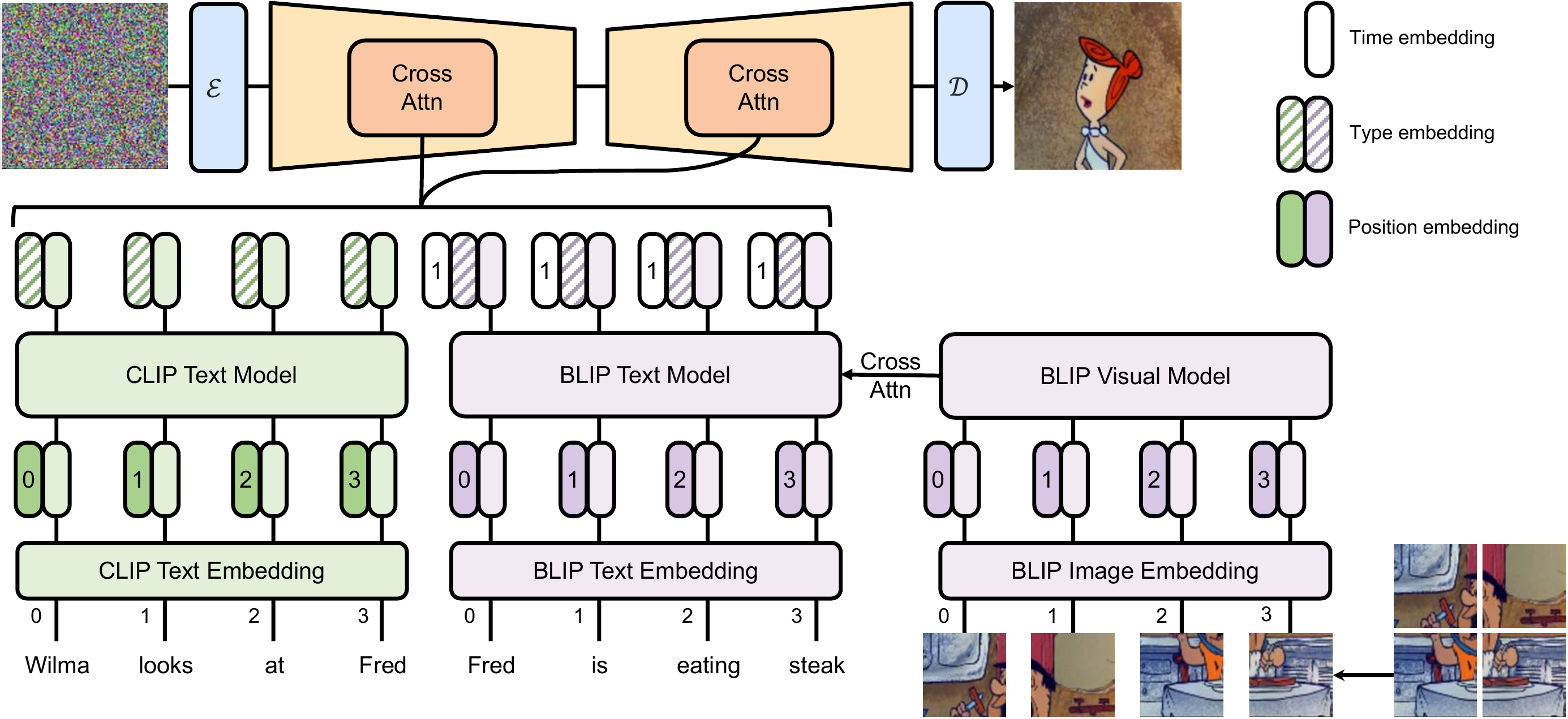}
\caption{Model Architecture.}
\label{fig:arch}
\end{subfigure}
\caption{Overview of proposed AR-LDM. The blue blocks represent the perceptual compression models; the orange blocks denote the generative network. The green blocks and the purple blocks are the history-aware conditioning network. Illustration inspired by~\cite{vilt}.}
\label{fig:modeloverview}
\end{figure*}

\subsection{Story Synthesis}
StoryGAN~\cite{storygan} firstly propose the story visualization task. Most story visualization models are based on GANs and comprise context text encoder, image generator, separate image, and story discriminators. The context text encoder and story discriminators mainly aim to preserve image consistency. DUCO-StoryGAN~\cite{ducostorygan} uses dual learning and copy-transform to improve story visualization. The copy-transform mechanism first incorporates features from previously generated images through attention mechanism to improve consistency. VLC-StoryGAN~\cite{vlcstorygan} and Word-Level SV~\cite{wordlevelsv} focus on text inputs, and propose to use structured input and sentence representation to better guide visual story generation. VP-CSV~\cite{vpcsv} leverages transformer~\cite{transformer}, and VQ-VAE~\cite{vqvae} to preserve characters across generated images. StoryDALL·E~\cite{storydalle} turns to story continuation, a variant of story visualization based on a given source image. They retrofit the pre-trained transformers DALL·E~\cite{dalle} and achieve a drastic improvement over GAN-based models.

\section{Method}
\subsection{Preliminaries}
Diffusion models~\cite{diffusionmodel} define a Markov chain of forward diffusion process $q$ to gradually add Gaussian noise sampled to real data $\mathbf{z}_0 \sim q(\mathbf{z})$ in $T$ steps. In particular, $\mathbf{z}$ in this paper denotes latent representations instead of pixels. The forward process $q(\mathbf{z}_t \vert \mathbf{z}_{t-1})$ at each time step $t$ is:
\begin{equation}
\begin{aligned}
    q(\mathbf{z}_t \vert \mathbf{z}_{t-1}) &= \mathcal{N}(\mathbf{z}_t; \sqrt{1 - \beta_t} \mathbf{z}_{t-1}, \beta_t\mathbf{I}) \quad\\
    q(\mathbf{z}_{1:T} \vert \mathbf{z}_0) &= \prod^T_{t=1} q(\mathbf{z}_t \vert \mathbf{z}_{t-1})
\end{aligned}
\end{equation}
in which $\beta_t \in (0, 1)$ denotes the step size. Note $\beta_{t-1} < \beta_t$.

Diffusion models learn a UNet~\cite{unet} denoted as $\boldsymbol{\epsilon}_{\theta}$ to reverse the forward diffusion process, constructing desired data samples from the noise. Let $\alpha_t = 1-\beta_t$ and $\bar{\alpha}_t = \prod_{i=1}^t \alpha_i$. We can reparameterize the denoising process $p(\mathbf{z}_{t-1} \vert \mathbf{z}_t)$ also as a Gaussian distribution. It can be estimated by $\boldsymbol{\epsilon}_{\theta}$ and has a form of the following:
\begin{equation}
\begin{aligned}
p_\theta(\mathbf{z}_{t-1} \vert \mathbf{z}_t) &= \mathcal{N}(\mathbf{z}_{t-1}; \boldsymbol{\mu}_\theta(\mathbf{z}_t, t), \boldsymbol{\Sigma}_\theta(\mathbf{z}_t, t))  \\
\text{with} \quad \boldsymbol{\mu}_\theta(\mathbf{z}_t, t) &= \frac{1}{\sqrt{\alpha_t}} (\mathbf{z}_t - \frac{\beta_t}{\sqrt{1 - \bar{\alpha_t}}}\boldsymbol{\epsilon}_{\theta}(\mathbf{z}_t, t) )
\end{aligned}
\end{equation}

The learning objective of diffusion models is to approximate the mean $\boldsymbol{\mu}_\theta(\mathbf{z}_t, t)$ in the reverse diffusion process. We can use variational lower bound (ELBO)~\cite{vae} to minimize the negative log-likelihood of $p_\theta(\mathbf{z}_{0})$~\cite{ddpm}, the simplified objective can be written as a denoising objective:
\begin{equation}
\mathcal{L} = \mathbb{E}_{\mathbf{z}_0, \boldsymbol{\epsilon} \sim \mathcal{N}(0, 1), t} \Big[\|\boldsymbol{\epsilon} - \boldsymbol{\epsilon}_\theta(\mathbf{z}_t, t)\|^2 \Big]
\label{eq:objective}
\end{equation}

During inference, \cite{classifierfree} proposes to use classifier-free guidance to obtain more relevant generation results.
\begin{equation}
    \hat{\boldsymbol{\epsilon}} = w\cdot \boldsymbol{\epsilon}_{\theta}(\mathbf{z}_t, \varphi, t)-(w-1)\cdot\boldsymbol{\epsilon}_{\theta}(\mathbf{z}_t, t)
\end{equation}
where $w$ is guidance scale, $\varphi$ denotes the condition.

\subsection{Auto-Regressive Latent Diffusion Model}
As we discussed in \cref{sec:introduction}, different from single caption text-to-image task, synthesizing coherent stories requires the model to be aware of history descriptions and scenes. For instance, consider a story ``\textit{A red metallic cylinder cube is at the center. Then add a green rubber cube at the right.}'' present in~\cite{storygan}. The second sentence alone cannot give the model enough guidance to generate a coherent image. It is crucial for the model to understand the history caption, the scene, and the appearance of the ``\textit{red metallic cylinder cube}'' in the first generated image. The key point of designing a strong story synthesis model is to make it capable of incorporating history captions and scenes for current image generation.

In this work, we propose auto-regressive latent diffusion model (AR-LDM) to achieve better consistency across frames. As shown in \cref{fig:ar}, AR-LDM leverage the history captions and images for future frame generation. For a certain story with a length of $L$, let $\mathbf{C} = [\mathbf{c}_1,\cdots, \mathbf{c}_j, \cdots, \mathbf{c}_L]$ be input captions and $\mathbf{X} = [\mathbf{x}_1, \cdots, \mathbf{x}_j, \cdots,\mathbf{x}_L]$ be the image targets, each caption $\mathbf{c}_j$ is corresponding to an image $\mathbf{x}_j \in \mathbb{R}^{C\times H\times W}$. Existing works assume conditional independence between each frame and generate the whole visual story according to the captions. While AR-LDM gets rid of this assumption by additionally conditioned on history images $\hat{\mathbf{x}}_{<j}$ and directly estimating the posterior based on the chain rule, which has a form of

\begin{equation}
\begin{aligned}
    P_{\text{AR}}(\mathbf{X}|\mathbf{C}) &= \prod_{j=1}^L P(\mathbf{x}_j \vert \hat{\mathbf{x}}_{<j}, \mathbf{C})\\
    &= \prod_{j=1}^L P(\mathbf{x}_{j} \vert \tau_{\theta}(\hat{\mathbf{x}}_{<j}, \mathbf{c}_{\leq j}))\\
    &= \prod_{j=1}^L p_\theta(\mathbf{z}_{0}^{[j]} \vert \tau_{\theta}(\mathcal{D}(\mathbf{z}_0^{[<j]}), \mathbf{c}_{\leq j}))\\
\end{aligned}
\label{eq:ar}
\end{equation}
where $p_\theta$ is the reverse diffusion process reparameterized by the generative network $\boldsymbol{\epsilon}_{\theta}$, and $\tau_{\theta}$ denotes the history-aware conditioning network. $\mathcal{D}$ denotes the decoder of the perceptual compression model (i.e., VQ-VAE), which also contains an encoder $\mathcal{E}$. To avoid abuse of notations, we use $\mathbf{z}_t^{[j]}$ to denote the latent diffusion variable at $j$-th frame and $t$-th diffusion step. \cref{fig:arch} shows the detailed architecture of AR-LDM.

\paragraph{Generative network}
Following~\cite{ldm}, AR-LDM also performs the forward and reverse diffusion processes in an efficient, low-dimensional latent space. The latent space is approximately perceptually equivalent to high-dimensional RGB space, while the redundant semantically meaningless information in pixels is eliminated. Specifically, perceptual compression models consisting of $\mathcal{E}$ and $\mathcal{D}$ are trained to encode the real data into the latent space and reverse, such that $\mathcal{D}(\mathcal{E}(\mathbf{x})) \approx \mathbf{x}$. AR-LDM uses latent representations $\mathbf{z} = \mathcal{E}(\mathbf{x})$ instead of pixels during the diffusion process, the final output can be decoded back to pixel space with $D(\mathbf{z})$. The separate mild perceptual compression stage only eliminates imperceptible details, allowing the model to achieve competitive generation results with much lower training and inference cost.

\paragraph{History-Aware Conditioning Network}
We use a history-aware conditioning network to encode the history caption-image pairs into a multimodal condition $\varphi_j = \tau_{\theta}(\hat{\mathbf{x}}_{<j}, \mathbf{c}_{\leq j})$ to guide denoising process $p_\theta(\mathbf{z}_{t}^{[j]} \vert \varphi_j)$. The estimated noise in \cref{eq:objective} can be rewritten as $\boldsymbol{\epsilon}_\theta(\mathbf{z}_t^{[j]}, \varphi_j, t)$. Therefore, $P(\mathbf{x}_j \vert \hat{\mathbf{x}}_{<j}, \mathbf{C})$ in \cref{eq:ar} can be simplified as $p_\theta(\mathbf{z}_{0}^{[j]} \vert \varphi_j)$. The conditioning network consists of CLIP~\cite{clip} and BLIP~\cite{blip}, in charge of current caption encoding and previous caption-image encoding, respectively. BLIP is pre-trained using vision-language understanding and generation tasks with large-scale filtered clean web data. Compared to CLIP which simply concatenates unimodal embeddings, BLIP utilizes the cross-attention module to deeply integrate visual and language modalities. It is able to ground the entities generated in history frames, allowing the generative network to refer to history scenes.

In summary, AR-LDM can generate image $\hat{\mathbf{x}}_{j}$ through:
\begin{equation}
\begin{aligned}
    \overline{\mathbf{c}}_j &= \mathrm{CLIP}(\mathbf{c}_j)\\
    \overline{\mathbf{m}}_{<j} &= 
    \begin{bmatrix}
    \mathrm{BLIP}(\mathbf{c}_1, \hat{\mathbf{x}}_{1}); \cdots; \mathrm{BLIP}(\mathbf{c}_{j-1}, \hat{\mathbf{x}}_{j-1})]
    \end{bmatrix}\\
    \varphi_j &= 
    \begin{bmatrix}
        \overline{\mathbf{c}}_j+\mathbf{c}^{type}; \overline{\mathbf{m}}_{<j}+\mathbf{m}^{type}+\mathbf{m}_{<j}^{time}
    \end{bmatrix}\\
    \mathbf{z}^{[j]}_0 &\sim p_\theta(\mathbf{z}^{[j]}_{0} \vert \varphi_j)\\
    \hat{\mathbf{x}}_{j} &= \mathcal{D}(\mathbf{z}^{[j]}_0)
\end{aligned}
\end{equation}
where $\overline{\mathbf{m}}_{<j}$ denotes encoded multimodal features from previous captions and generated images. $\mathbf{c}^{type}, \mathbf{m}^{type} \in \mathbb{R}^{D}$ are text and multimodal type embedding, respectively. $D=768$ denotes the embedding dimension. $\mathbf{m}^{time} \in \mathbb{R}^{L\times D}$ is time embedding. Specifically, the first image $\mathbf{x}_1$ is provided as input for the story continuation setting.

\subsection{Adaptive AR-LDM}
\label{sec:unseen_character_adaptation}
For real-world applications like comic drawing, it's necessary to preserve consistency for the new (unseen) characters. As we discussed in \cref{sec:introduction}, it is challenging because one cannot depict every single detail of the unseen character in captions, and the story synthesis model always suffers from the inconsistent descriptions of a certain unseen character like the generated results of AR-LDM shown in \cref{fig:textualinversion}. Inspired by Textual Inversion~\cite{textualinversion} and DreamBooth~\cite{dreambooth}, we add a new token \texttt{<char>} to represent the unseen character, and adapt the trained AR-LDM to generalize to the specific unseen character. Specifically, the embedding of the new token \texttt{<char>} is initialized by that of a similar existing word, like ``\textit{man}'' or ``\textit{woman}''. We need only 4-5 images of the character to compose a story as our training dataset, and finetuned AR-LDM for 100 epochs using the same learning rate of 1e-5. We find that finetuning whole parameters of AR-LDM (only except the encoder $\mathcal{E}$ and decoder $\mathcal{D}$) results in a better performance.

\begin{figure}[!t]
    \centering
    \includegraphics[width=\linewidth]{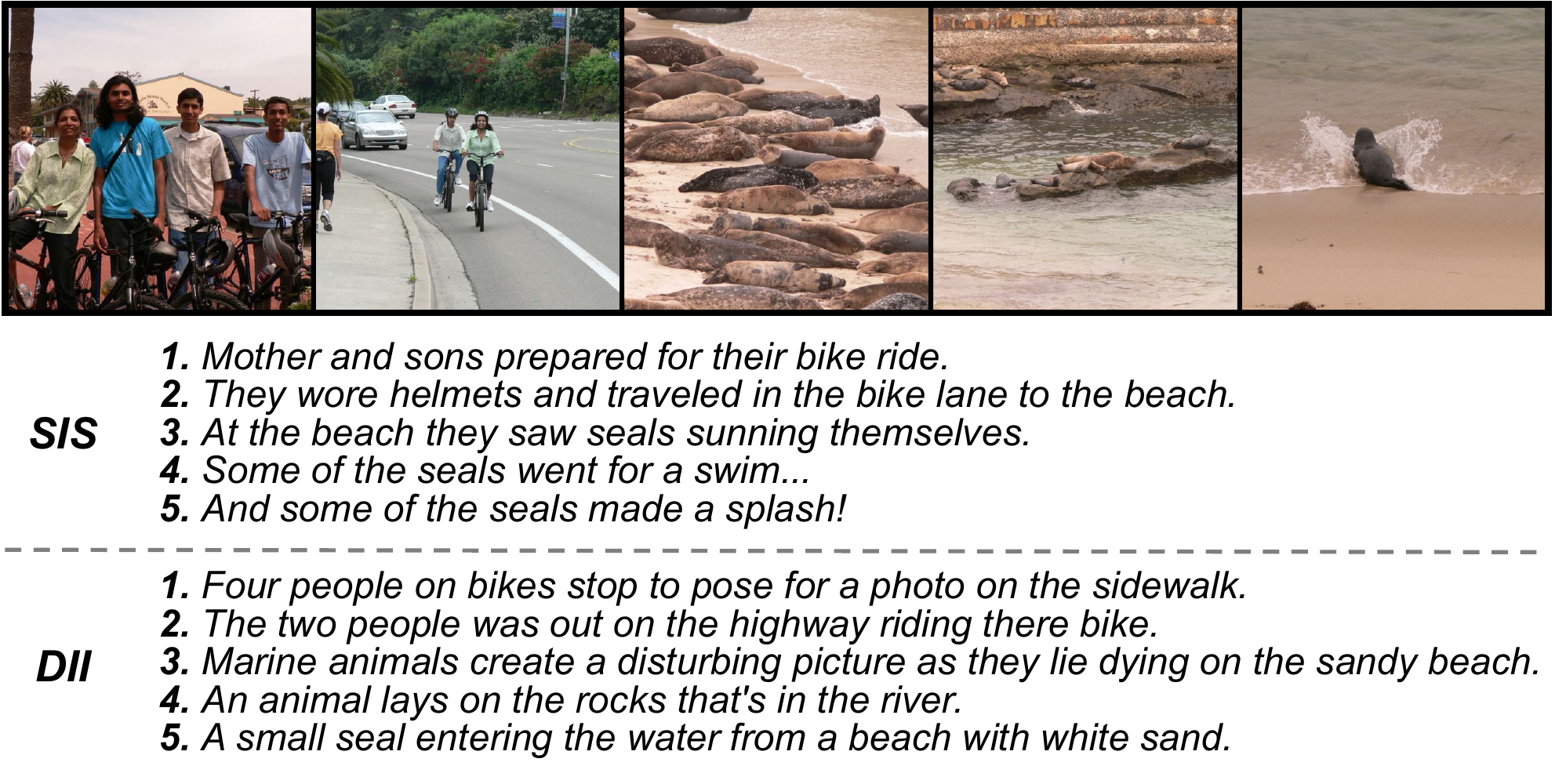}
    \caption{A data sample from VIST with description-in-isolation (DII), and story-in-sequence (SIS) captions.}
    \label{fig:vistsamples}
\end{figure}

\begin{table*}[t]
\small
\centering
\setlength\tabcolsep{12pt}
\begin{tabularx}{\linewidth}{cccccc}
    \toprule
    Models & \# of Params & PororoSV & FlintstonesSV & VIST-SIS & VIST-DII \\
    \midrule
    StoryGANc~\cite{storydalle} & - & 74.63 & 90.29 & - & - \\
    StoryDALL·E (prompt tuning)~\cite{storydalle} & 1.3B & 61.23 & 53.71 & - & - \\
    StoryDALL·E~\cite{storydalle} & 1.3B & 25.90 & 26.49 & - & - \\
    MEGA-StoryDALL·E~\cite{storydalle} & 2.8B & 23.48 & 23.58 & 20.98$^{*}$ & 24.61$^{*}$ \\
    AR-LDM (Ours) & 1.5B & \textbf{17.40} & \textbf{19.28} & \textbf{16.95} & \textbf{17.03} \\
    \bottomrule
    \end{tabularx}
\caption{Story continuation FID scores (lower is better) of AR-LDM and several previous models. $^{*}$ denotes experimental results reproduced by us, where we trained MEGA-StoryDALL·E for 50 epochs using the same training strategies as AR-LDM.}
\label{tb:continuationfid}
\vspace{-8pt}
\end{table*}

\section{Experiments}
\subsection{Datasets}
We use three datasets as our testbed, PororoSV~\cite{storygan}, FlintstonesSV~\cite{vlcstorygan}, and VIST~\cite{vist}. Each story in these three datasets contains 5 consecutive frames. For story visualization, we predict all 5 frames from captions. For story continuation, the first frame is assigned as a source frame, and we generate the rest 4 frames with reference to the source frame. In this section, we will briefly go through these three datasets. For a more detailed introduction, see \cref{sec:detail_of_used_datasets}.

\paragraph{PororoSV and FlintstonesSV}
The PororoSV~\cite{storygan} and FlintstonesSV~\cite{vlcstorygan} datasets are adapted from Pororo video question answering dataset~\cite{pororo} and Flintstones text-to-video synthesis dataset~\cite{flintstones}, respectively. Both of the two datasets contain several recurring characters. While FlintstonesSV is relatively harder than PororoSV for there are many unseen characters within the stories.

\paragraph{VIST}
However, there are two major limitations of existing story synthesis datasets: (1) current datasets are all cartoon ones; (2) sentences are isolated descriptions rather than sequential stories. We propose to use the Visual Story Telling (VIST) dataset~\cite{vist} for real-world story synthesizing. VIST provides two kinds of captions: description-in-isolation (DII) and story-in-sequence (SIS). As shown in \cref{fig:vistsamples}, DII captions are more like the ones in PororoSV and FlintstonesSV, every single caption contains detailed information about the image. In contrast, SIS captions describe the five images like a story and merely repeat the content mentioned before. The story-style captions are closer to real-world use cases and require the model to have a better contextual understanding ability.

\subsection{Experimental Settings}
Our model is initialized by the weight of stable diffusion~\cite{ldm}, a publicly available text-to-image latent diffusion model trained on LAION-5B~\cite{laion}. We trained AR-LDM for 50 epochs on 8 NVIDIA A100-80GB GPUs for two days. We only freeze the encoder $\mathcal{E}$ and decoder $\mathcal{D}$, and finetune the rest parameters using the AdamW optimizer~\cite{adamw} with an initial learning rate of $1 \times 10^{-5}$ and a weight decay of $10^{-4}$. During training, we randomly drop the condition $\varphi$ at a probability of 0.1 for each frame. A cosine scheduler and 8000 steps linear learning rate warm-up are used during training. During inference, we sample images using the DDIM scheduler~\cite{ddim} for 250 inference steps with guidance scale $w$ set to 6.0.

\begin{table}[!t]
\small
\centering
\setlength\tabcolsep{30pt}
\begin{tabularx}{\linewidth}{cc}
    \toprule
    Models & FID \\
    \midrule
    StoryGAN~\cite{storygan} & 158.06  \\
    CP-CSV~\cite{cpcsv} & 149.29 \\
    DUCO-StoryGAN~\cite{ducostorygan} & 96.51  \\
    VLC-StoryGAN~\cite{vlcstorygan} & 84.96  \\
    VP-CSV~\cite{vpcsv} & 65.51 \\
    Word-Level SV~\cite{wordlevelsv} & 56.08 \\
    AR-LDM (Ours) & \textbf{16.59} \\
    \bottomrule
    \end{tabularx}
\caption{Story visualization FID score results on PororoSV. We use the results reported by~\cite{vpcsv} and~\cite{wordlevelsv}.}
\label{tb:visualizationfid}
\end{table}

\begin{figure*}[!th]
\centering
\begin{subfigure}{\linewidth}
    \centering
    \includegraphics[width=\linewidth]{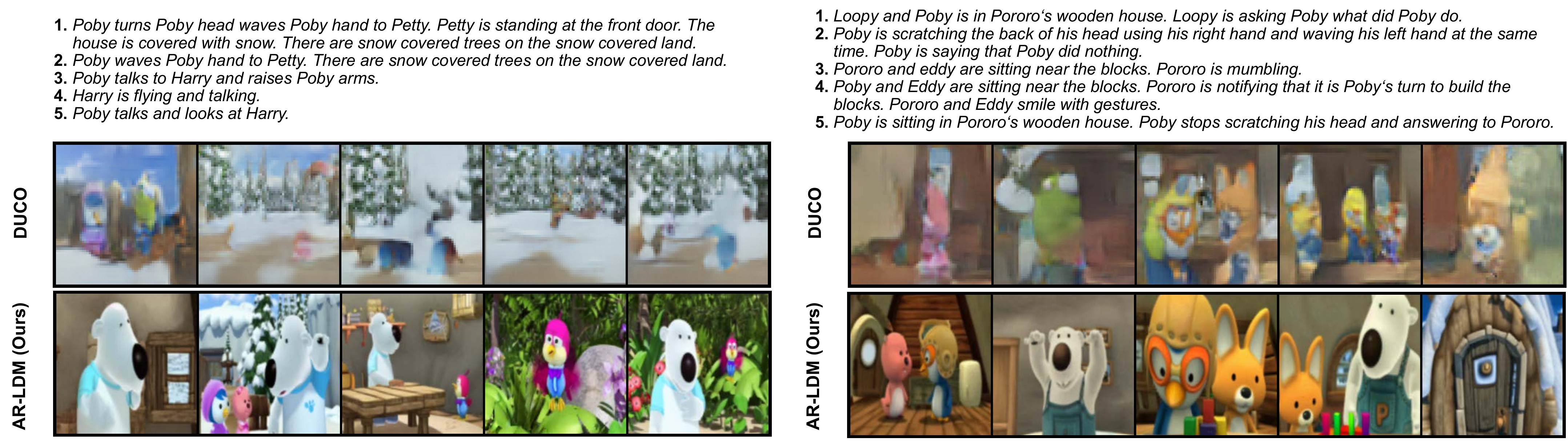}
    \caption{Comparison of \textbf{story visualization} results between AR-LDM and DUCO-StoryGAN~\cite{ducostorygan}. DUCO-StoryGAN also incorporates features of previously generated images through copy-transform, but we can observe that AR-LDM can faithfully generate high-quality images exactly as the captions described.}
    \label{fig:visualization}
\end{subfigure}
\begin{subfigure}{\linewidth}
    \centering
    \includegraphics[width=\linewidth]{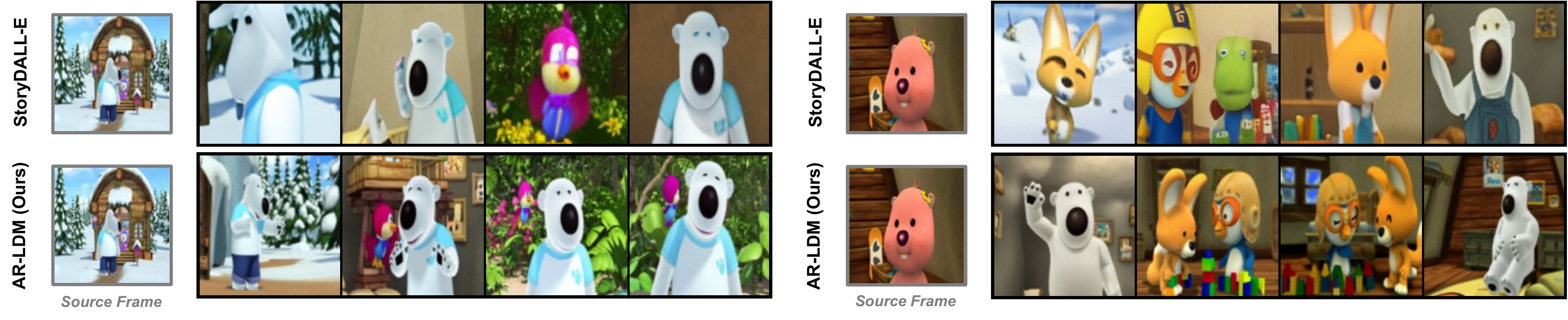}
    \caption{Comparison of \textbf{story continuation} results between AR-LDM and StoryDALL·E.~\cite{storydalle}}
    \label{fig:pororocontinuation}
\end{subfigure}
\caption{Visual story synthesis results on PororoSV. Note the case in \cref{fig:visualization} and \cref{fig:pororocontinuation} is the same one.}
\label{fig:pororovisandcontinue}
\end{figure*}

\begin{figure*}[!th]
    \centering
    \includegraphics[width=\linewidth]{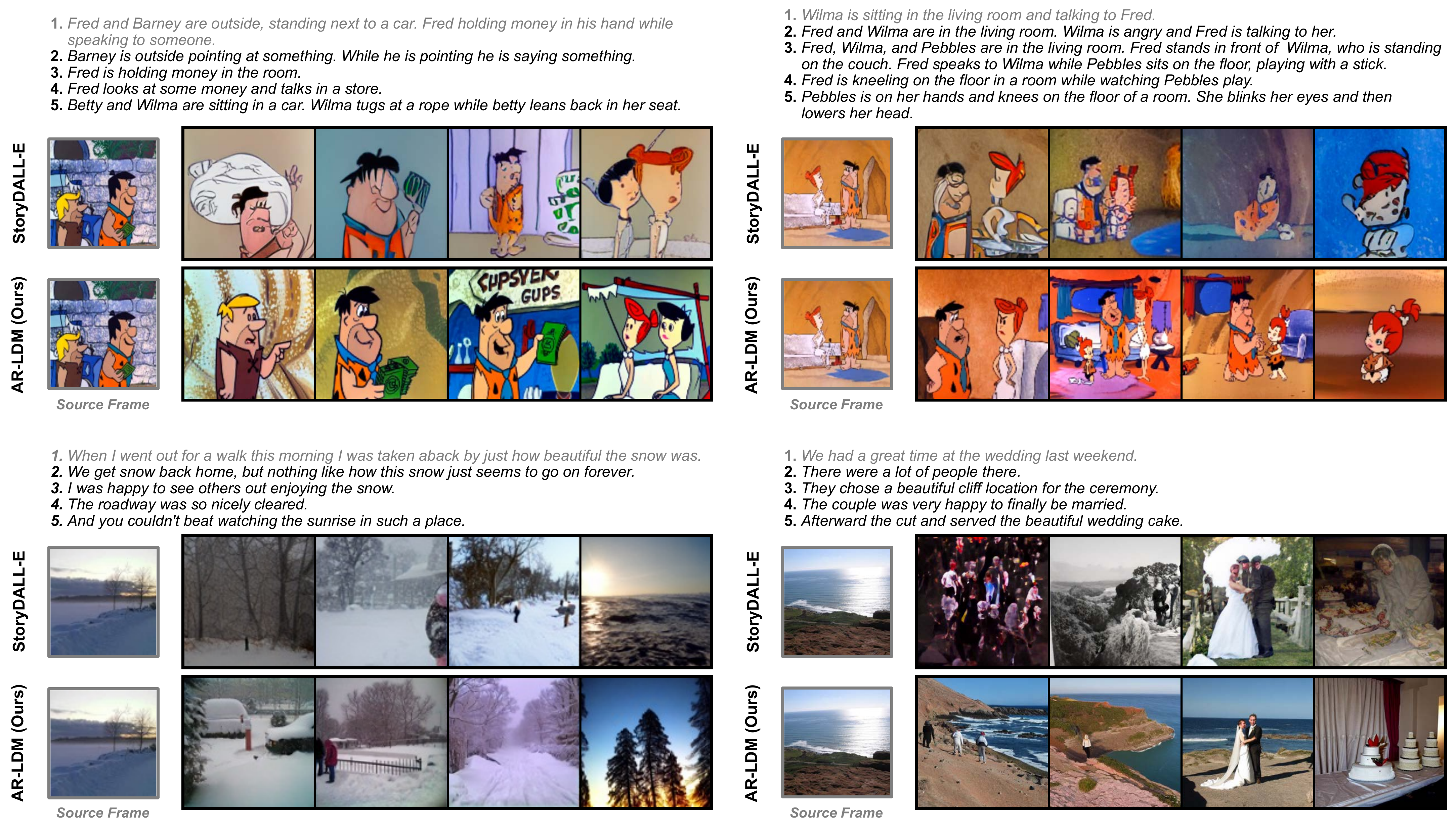}
    \caption{Comparison of story continuation results between AR-LDM and StoryDALL·E on FlintstonesSV (upper) and VIST-SIS (lower). Better visual quality, relevance, and consistency can be observed in the visual stories synthesized by AR-LDM.}
    \label{fig:continuation}
\end{figure*}

\section{Results}
We evaluate AR-LDM using two settings: (1) quantitative evaluation using automatic metric FID score~\cite{fid}; (2) large-scale human evaluations regarding visual quality, relevance, and consistency. Note that we report FID scores at a resolution of 64$\times$64 following~\cite{storydalle} for a fair comparison.

\subsection{Quantitative Results}
\paragraph{Story Visualization}
In \cref{tb:visualizationfid}, we present our story visualization results on PororoSV, which is the most popular benchmark used by previous works. AR-LDM makes significant progress and achieves a SoTA FID score of 16.59, lagging previous methods by a large margin. This suggests the superior visual quality of AR-LDM's generated visual stories. As shown in \cref{fig:visualization}, AR-LDM is able to generate high-quality, coherent visual stories while faithfully reproducing character details and backgrounds. Additional cases can be found in \cref{sec:additional_pororo}.

\begin{table}[!t]
\small
\centering
\setlength\tabcolsep{11pt}
\begin{tabularx}{\linewidth}{lcc}
    \toprule
    Models & FID & $\Delta$\\
    \midrule
    Stable Diffusion~\cite{ldm} & 22.10 & - \\
    \quad + CLIP Visual Conditioning & 21.01 & -1.09\\
    \midrule
    Stable Diffusion~\cite{ldm} & 22.10 & - \\
    \quad + BLIP Multimodal Conditioning & 19.94 & -2.16 \\
    \quad \quad + Auto Regression & \textbf{19.28} & -2.82 \\
    \bottomrule
    \end{tabularx}
\caption{Ablation study results for story continuation task on FlintstonesSV. All models are finetuned using the same training data and strategies as AR-LDM.}
\label{tb:ablation}
\end{table}

\begin{figure}[!t]
    \centering
    \includegraphics[width=\linewidth]{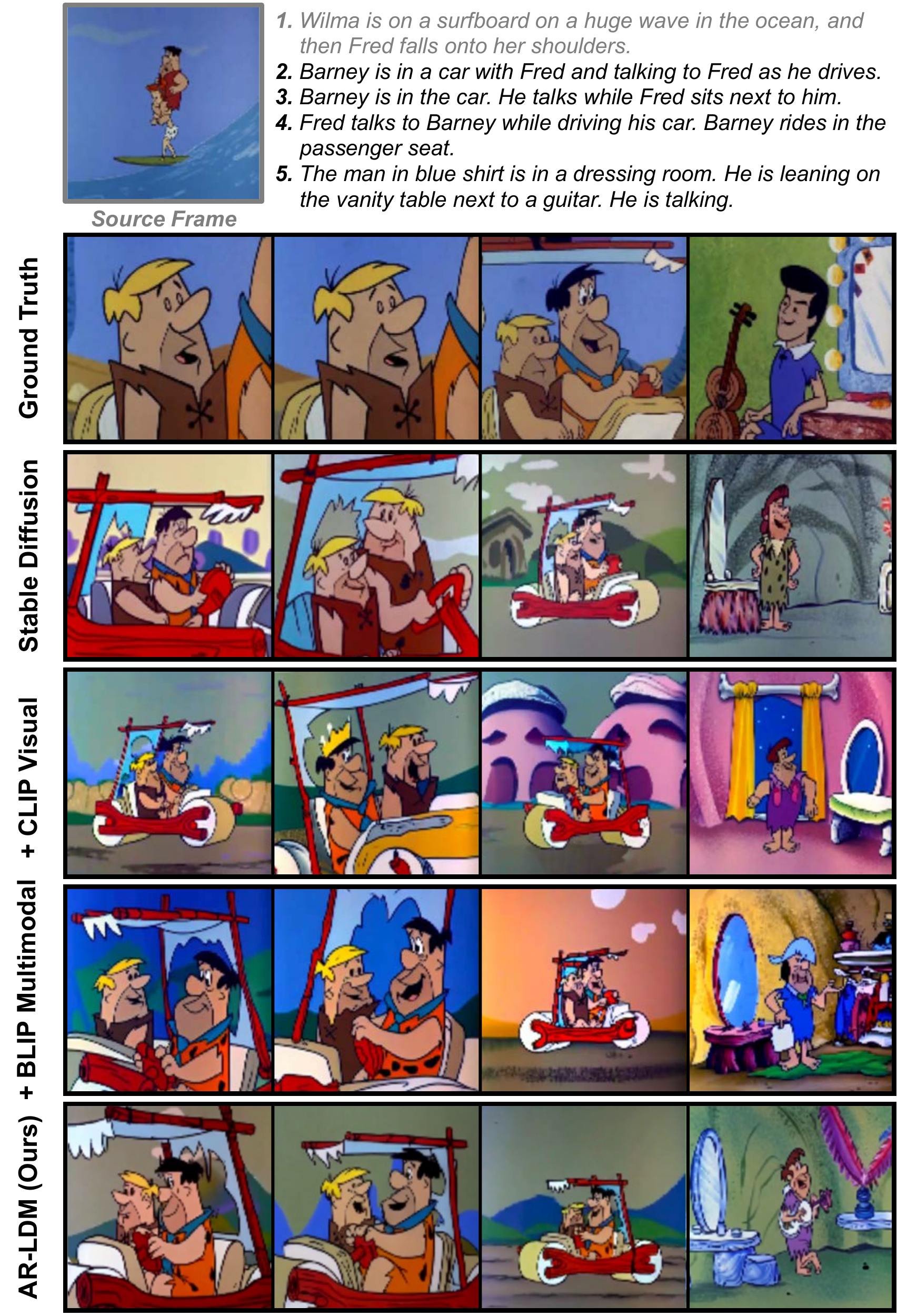}
    \caption{Synthesized visual story examples for different ablation models mentioned in \cref{tb:ablation}}
    \label{fig:ablation}
\end{figure}

\paragraph{Story Continuation}
We test the story continuation performance of AR-LDM and present the results in \cref{tb:continuationfid}. AR-LDM achieves a series of new SoTA FID scores on all four datasets.
It is worth mentioning that AR-LDM outperforms MEGA-StoryDALL·E with approximately half parameters. As shown in \cref{fig:pororocontinuation}, AR-LDM can preserve scenes through auto-regression generation, like the background of the last two frames in the left case, as well as the blocks in the third and fourth frames in the right case. \cref{fig:continuation} show more cases on FlintstonesSV and VIST-SIS datasets, we can observe scenes consistency across frames, for example, the windows in the third and fourth frames of the upper left case, and the coastal scene in the lower left case. Moreover, we can observe that AR-LDM is able to infer from previous captions and scenes. Taking the lower left case as an example, AR-LDM generates a cleared snowy roadway in the third frame without relying on a caption with description of snowy weather. The last caption does not directly describe what ``\textit{such a place}'' looks like. 
AR-LDM generates a scene of trees occurring in the previously generated image, which is more reasonable than the results of StoryDALL·E. Additional cases can be found in \cref{sec:additional_synthesized_visual_stories}.

\paragraph{Ablation Studies}
We conduct ablation studies on the proposed auto-regressive multimodal module. Particularly, we use the story continuation task on FlintstonesSV as our test bed, for it requires higher consistency across frames and contains many challenging unseen characters. Starting from a finetuned Stable Diffusion baseline, we utilize a source frame image to guide the generation through CLIP visual encoder and obtain an FID score improvement of 1.09. If we further leverage the source frame image caption pair and use BLIP to encode it into a multimodal embedding to condition the diffusion process, we can observe an improvement of 2.16 compared to the baseline model. Finally, we employ an auto-regressive generation manner and achieve a further improvement of 0.66. However, FID score is only related to visual quality. Apart from visual quality, AR-LDM also performs better in terms of relevance and consistency, we provide a case in FlintstonesSV to illustrate it. As shown in \cref{fig:ablation}, compared to other methods, AR-LDM with auto-regressive generation manner better preserves backgrounds and view of the scene across frames.

\begin{table}[t]
\small
\centering
\setlength\tabcolsep{4pt}
\begin{tabularx}{\linewidth}{c|c|ccc}
    \toprule
    Dataset & Criterion & Win (\%) & Tie (\%) & Lose (\%) \\
    \hline
    \multirow{3}*{PororoSV} & Visual Quality & \textbf{90.6} & 2.2 & 7.2 \\
    ~ & Relevance & \textbf{88.6} & 1.0 & 10.4 \\
    ~ & Consistency & \textbf{70.6} & 9.4 & 20.0 \\
    \hline
    \multirow{3}*{FlintstonesSV} & Visual Quality & \textbf{99.4} & 0.2 & 0.4 \\
    ~ & Relevance & \textbf{96.2} & 1.0 & 2.8\\
    ~ & Consistency & \textbf{93.2} & 6.0 & 0.8 \\
    \hline
    \multirow{3}*{VIST-SIS} & Visual Quality & \textbf{86.6} & 6.6 & 6.8 \\
    ~ & Relevance & \textbf{69.8} & 6.8 & 23.4 \\
    ~ & Consistency & \textbf{81.6} & 8.2 & 10.2 \\
    \bottomrule
    \end{tabularx}
\caption{Human evaluation results of story continuation task on PororoSV, FlintstonesSV, and VIST-SIS datasets. Win means AR-LDM is preferred over StoryDALL·E, Lose for vice-versa, Tie denotes the samples that human annotators can hardly choose.}
\label{tb:humaneval}
\vspace{-8pt}
\end{table}

\subsection{Large-Scale Human Evaluation}
\label{sec:large_scale_human_evaluation}
Apart from quantitative results, we also carry out large-scale human evaluations for the story continuation task on PororoSV, FlintstonesSV, and VIST-SIS datasets in terms of visual quality, relevance, and consistency. It is necessary because our quantitative metric, the FID score, only measures visual quality. What's more, Inception-V3~\cite{inception} backbone may also cause a mismatch between the FID score and human preference for its locality inductive bias and spatial translation invariance~\cite{inductivebias}.

The annotation group compares the synthesized stories of AR-LDM vs. the ones of StoryDALL·E. We randomly select 500 samples of each dataset to be evaluated for each criterion. The evaluation results are shown in \cref{tb:humaneval}. For detailed human evaluation settings, see \cref{sec:detailed_human_evaluation_settings}. Owing to the powerful latent diffusion model, AR-LDM significantly outperforms StoryDALL·E in visual quality. The history-aware conditioning network also largely boosts AR-LDM's performance in terms of relevance and consistency. Examples of win or lose cases are shown in \cref{sec:win_and_lose_cases_in_human_evaluation}.

\begin{figure}[!t]
    \centering
    \includegraphics[width=\linewidth]{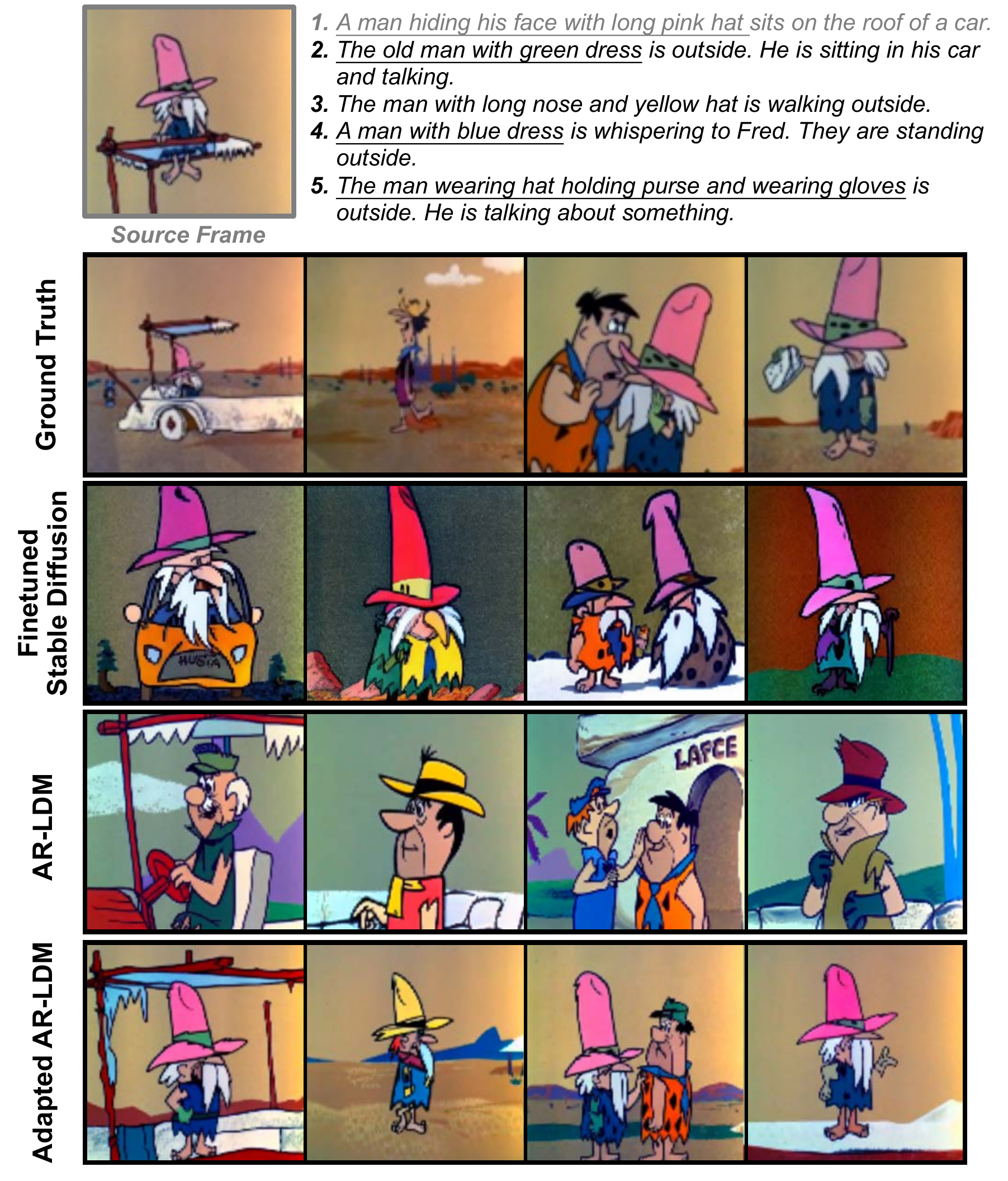}
    \caption{Adaptation results for a case AR-LDM failed to properly generate on FlintstonesSV. The underlined texts refer to one specific person and can be replaced by \texttt{<char>} in adapted AR-LDM.}
    \label{fig:textualinversion}
\end{figure}

\subsection{Adapting to Unseen Characters}
\label{sec:adapting_to_unseen_characters}
As shown in \cref{fig:textualinversion}, all underlined texts are referring the same character (i.e., the man with a pink hat in the source frame), while the description is inconsistent. As a result, AR-LDM generates three different characters according to every single description. After being finetuned on 3-5 images, our adapted AR-LDM can generate consistent characters as well as faithfully synthesize scenes and characters as what captions describe. In contrast, simply finetune stable diffusion using same 3-5 images cannnot obtain satisfying generation result, because it confuses other characters with \texttt{<char>} and fails to generate them. Additional cases can be found in \cref{sec:additional_unseen_character_adaptation_results}.

\begin{table}[t]
\small
\centering
\setlength\tabcolsep{4pt}
\begin{tabularx}{\linewidth}{c|c|ccc}
    \toprule
    Dataset & Criterion & Win (\%) & Tie (\%) & Lose (\%) \\
    \hline
    \multirow{3}*{PororoSV} & Visual Quality & 41.8 & 17.4 & 40.8 \\
    ~ & Relevance & 18.0 & 28.6 & 53.4 \\
    ~ & Consistency & 3.8 & 3.2 & 93.0 \\
    \hline
    \multirow{3}*{FlintstonesSV} & Visual Quality & 42.2 & 20.0 & 37.8 \\
    ~ & Relevance & 24.6 & 26.4 & 49.0 \\
    ~ & Consistency & 2.6 & 13.2 & 84.2 \\
    \hline
    \multirow{3}*{VIST-SIS} & Visual Quality & 14.6 & 20.6 & 64.8 \\
    ~ & Relevance & 19.2 & 48.6 & 32.2 \\
    ~ & Consistency & 3.0 & 46.2 & 50.8 \\
    \bottomrule
    \end{tabularx}
\caption{Human evaluation results of story continuation task on PororoSV, FlintstonesSV, and VIST-SIS datasets. The comparison is between visual stories synthesized by AR-LDM and ground truth reference ones.}
\label{tb:limatition}
\vspace{-8pt}
\end{table}
\section{Limitations}
Though AR-LDM largely outperforms StoryDALL·E in the human evaluation as we discussed in \cref{sec:large_scale_human_evaluation}, we find that AR-LDM is still far behind ground truth visual stories in terms of consistency. As the human evaluation results shown in \cref{tb:limatition}, AR-LDM is comparable to ground truth visual stories regarding visual quality and relevance; We can also observe that 49.2\% of generated stories on VIST are as consistent as ground truth. However, as for more challenging PororoSV and FlintstonesSV datasets whose frames are sampled from videos, we find that few synthesized visual stories are as consistent as ground truth references. This indicates consistency is a short board of current generative models, and it still needs to be improved in the future. Additional discussion can be found in \cref{sec:additional_discussion}.

\section{Conclusion}
In this paper, we propose AR-LDM, the first work employing diffusion models for story visualization and continuation. AR-LDM incorporates captions and previously generated images into current frame generation through an auto-regressive manner. The powerful diffusion model backbone and the novel conditioning network demonstrate effectiveness in generating coherent visual stories in high quality. Meanwhile, we introduce an efficient strategy for AR-LDM to adapt to unseen characters. In addition, we introduce a real-world story synthesis dataset VIST with more challenging story-in-sequence captions. Quantitative results show that AR-LDM achieves remarkable improvements of FID score on PororoSV, FlintstonesSV, and VIST. Large-scale human evaluations show that human annotators prefer the generation of AR-LDM over previous methods. 
\clearpage
{\small
\bibliographystyle{ieee_fullname}
\bibliography{egbib}
}
\clearpage
\appendix
We regret that the ``Diffusion'' in the title is misspelled as ``Diffision'' in our original submission. We will correct this in the camera-ready version. We are sorry if this typo disturbed your reading.
\section{Details of Used Datasets}
\label{sec:detail_of_used_datasets}
\begin{figure}[h]
    \centering
    \includegraphics[width=\linewidth]{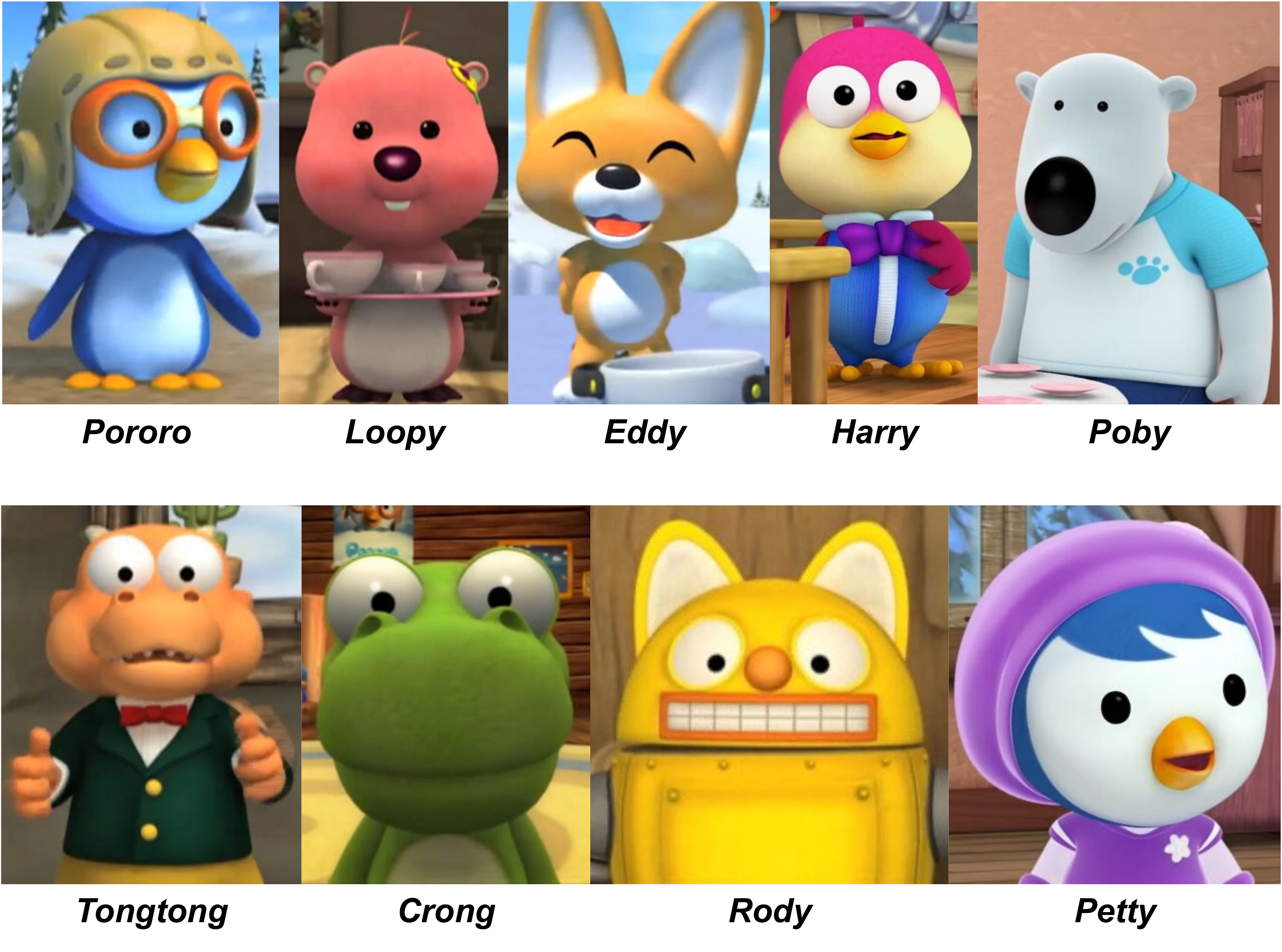}
    \caption{Main character names and corresponding photos in PororoSV. The photos are from \url{https://pororo.fandom.com/}}
    \label{fig:pororosv_char}
\end{figure}

\begin{figure}[h]
    \centering
    \includegraphics[width=\linewidth]{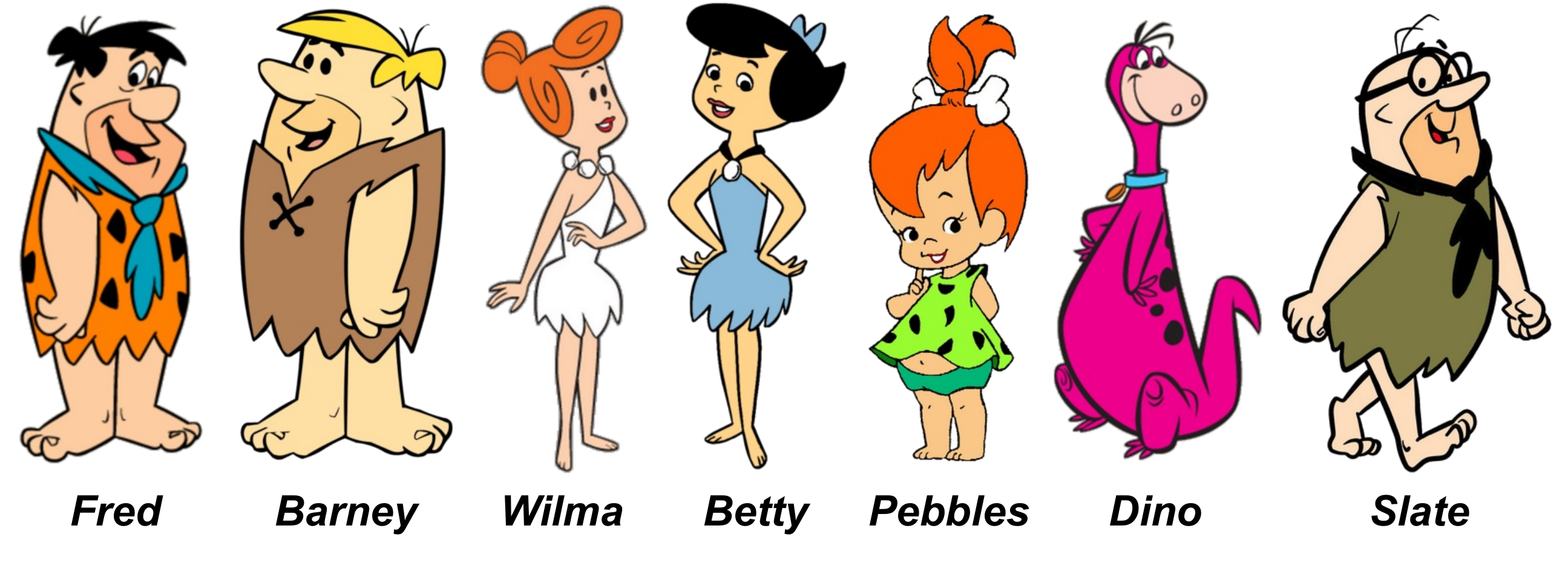}
    \caption{Main character names and corresponding photos in FlintstonesSV. The photos are from \url{https://flintstones.fandom.com/}}
    \label{fig:flintstonesv_char}
\end{figure}

\paragraph{PororoSV}
PororoSV contains 10191/2334/2208 samples of the train, val, and test set, respectively. Each sample contains 5 consecutive frames sampled from videos. There are 9 main characters in PororoSV: Pororo, Loopy, Eddy, Harry, Poby, Tongtong, Crong, Rody, and Petty. Profile pictures of them are given in \cref{fig:pororosv_char}.
\paragraph{FlintstonesSV}
FlintstonesSV contains 20132/2071/2309 samples of the train, val, and test set, respectively. Each sample contains 5 consecutive frames sampled from videos. There are 7 main characters in PororoSV: Fred, Barney, Wilma, Betty, Pebbles, Dino, and Slate. Profile pictures of them are given in \cref{fig:flintstonesv_char}.
\newpage
\paragraph{VIST}
VIST contains 23344/2921/2925 samples of the train, val, and test set, respectively. Each sample contains 5 consecutive frames and two kinds of captions: SIS and DII. There are no recurring characters in VIST. The VIST dataset used in this paper is a subset of the original one. Because some images are removed by their owners and are not accessible now, we drop the stories containing such images. We further choose stories that contain both SIS and DII captions.

\section{Detailed Human Evaluation Settings}
\label{sec:detailed_human_evaluation_settings}
We provide human evaluation results regarding visual quality, relevance, and consistency. Human annotators tend to choose visual stories in high visual quality. This may confuse the three separate evaluation criteria. To make criteria orthogonal to each other, we carefully design the human evaluation process. Specifically, we only provide single images (without captions) for visual quality evaluation, single images and the corresponding specific captions for relevance evaluation, and whole visual stories (without captions) for consistency evaluation. For the detailed annotation standards and instructions given to annotators, see \cref{sec:annotation_instructions_for_human_evaluation}.

\section{Win and Lose Cases in Human Evaluation}
\label{sec:win_and_lose_cases_in_human_evaluation}
\begin{figure*}[!h]
\begin{subfigure}{0.32\linewidth}
\centering
\includegraphics[height=0.93\textheight]{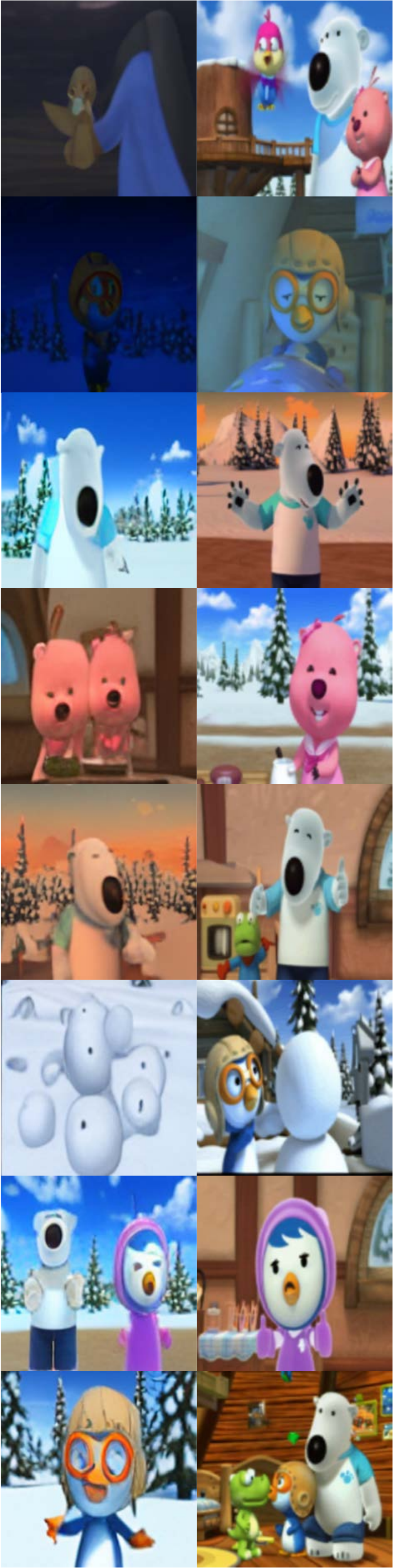}
\caption{Cases on PororoSV.}
\end{subfigure}
\hfill
\begin{subfigure}{0.32\linewidth}
\centering
\includegraphics[height=0.93\textheight]{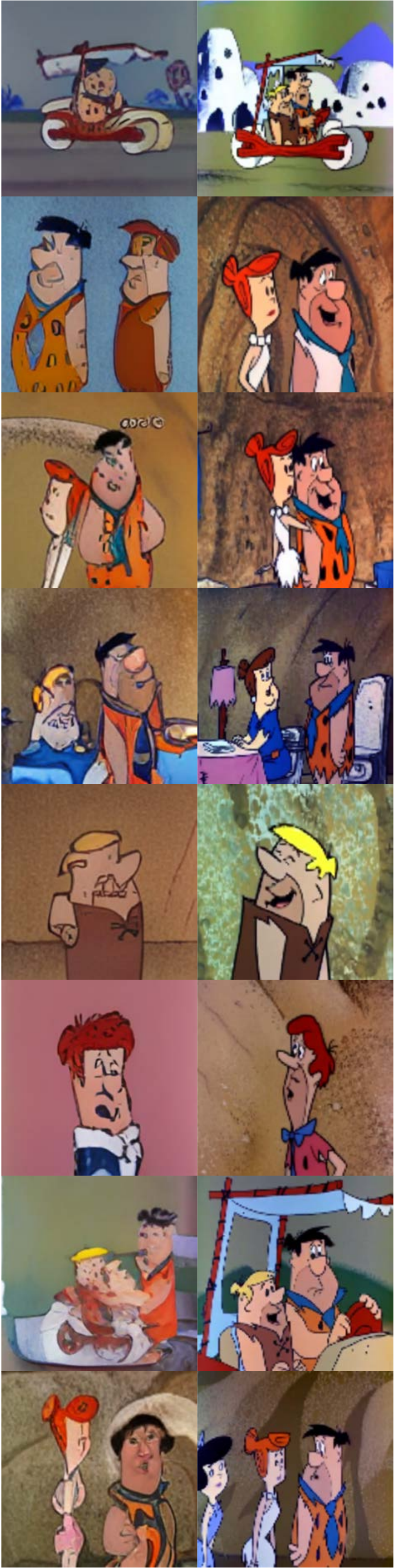}
\caption{Cases on FlintstonesSV.}
\end{subfigure}
\hfill
\begin{subfigure}{0.32\linewidth}
\centering
\includegraphics[height=0.93\textheight]{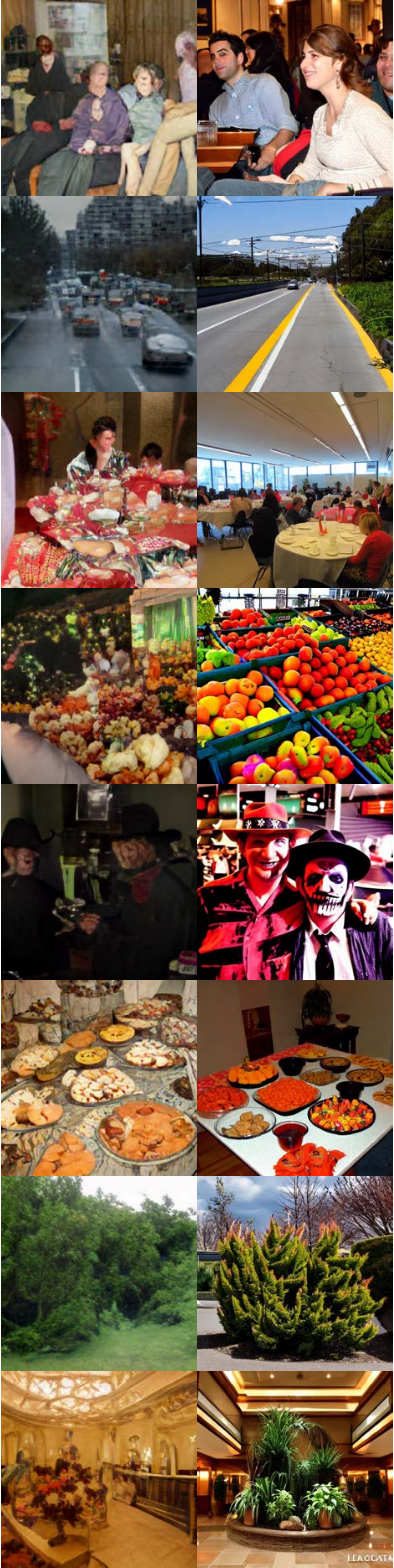}
\caption{Cases on VIST-SIS.}
\end{subfigure}
\caption{Cases that AR-LDM \textbf{wins} StoryDALL·E in human evaluation regarding \textbf{visual quality}. The left ones are synthesized by StoryDALL·E, and the right ones are synthesized by AR-LDM.}
\label{fig:win_quality}
\end{figure*}

\begin{figure*}[!h]
\begin{subfigure}{0.32\linewidth}
\centering
\includegraphics[height=0.93\textheight]{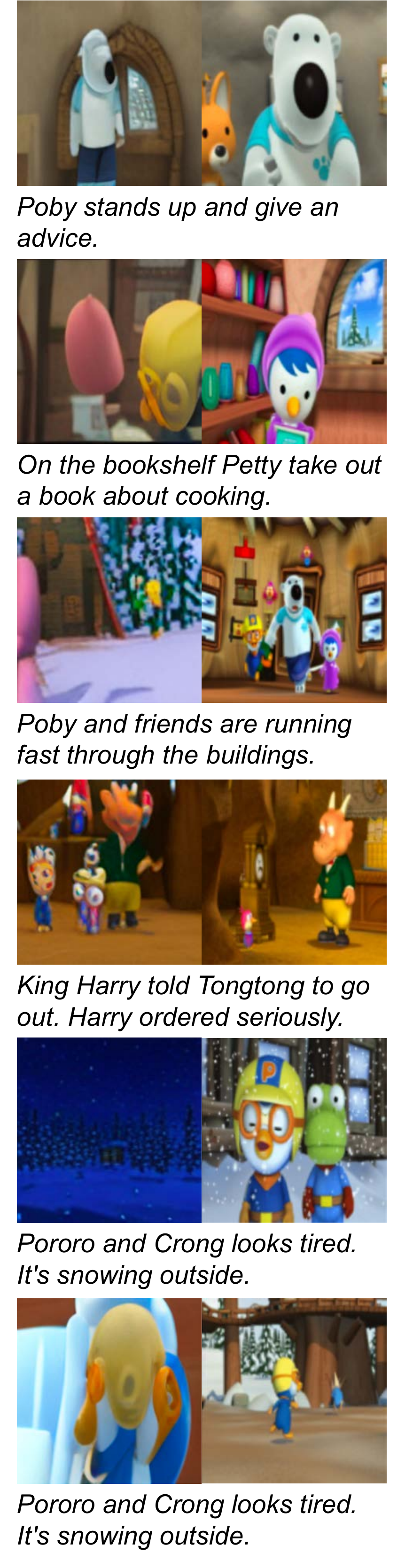}
\caption{Cases on PororoSV.}
\end{subfigure}
\hfill
\begin{subfigure}{0.32\linewidth}
\centering
\includegraphics[height=0.93\textheight]{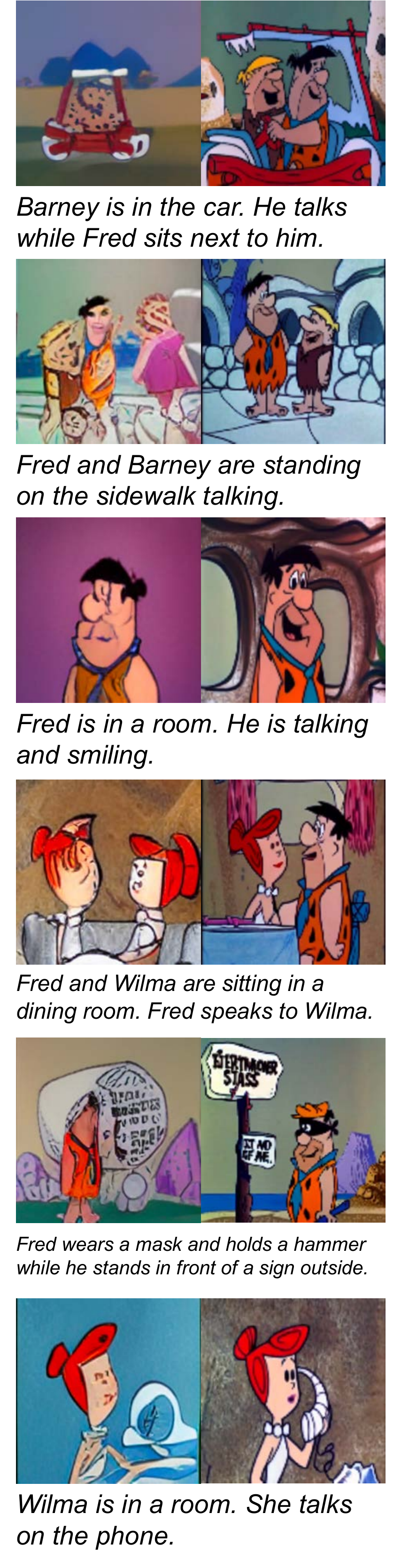}
\caption{Cases on FlintstonesSV.}
\end{subfigure}
\hfill
\begin{subfigure}{0.32\linewidth}
\centering
\includegraphics[height=0.93\textheight]{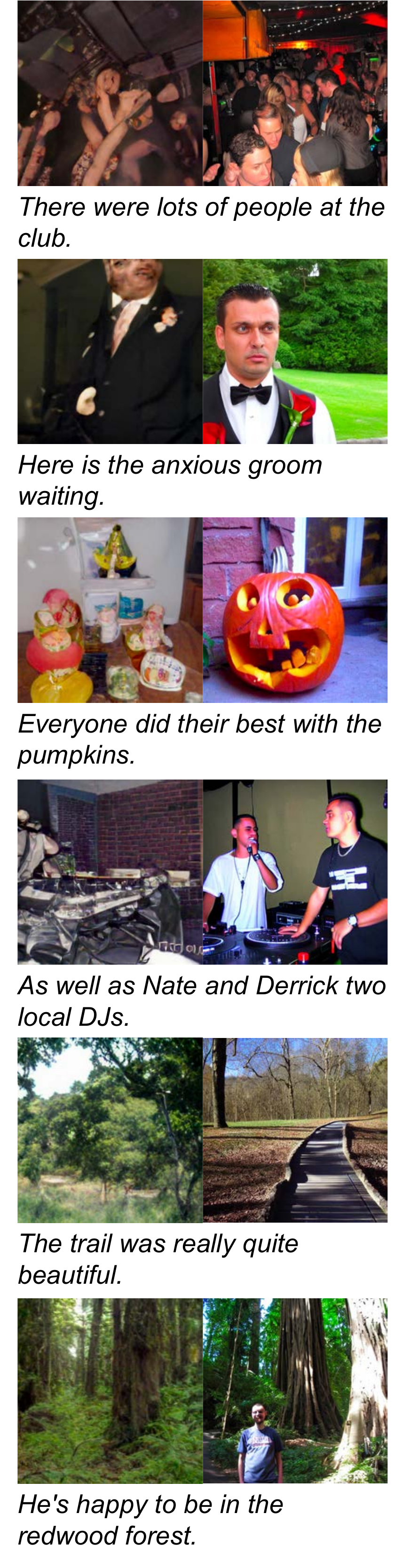}
\caption{Cases on VIST-SIS.}
\end{subfigure}
\caption{Cases that AR-LDM \textbf{wins} StoryDALL·E in human evaluation regarding \textbf{relevance}. The left ones are synthesized by StoryDALL·E, and the right ones are synthesized by AR-LDM.}
\label{fig:win_relevance}
\end{figure*}

\begin{figure*}[!h]
\centering
\begin{subfigure}{\linewidth}
\centering
\includegraphics[height=0.3\textheight]{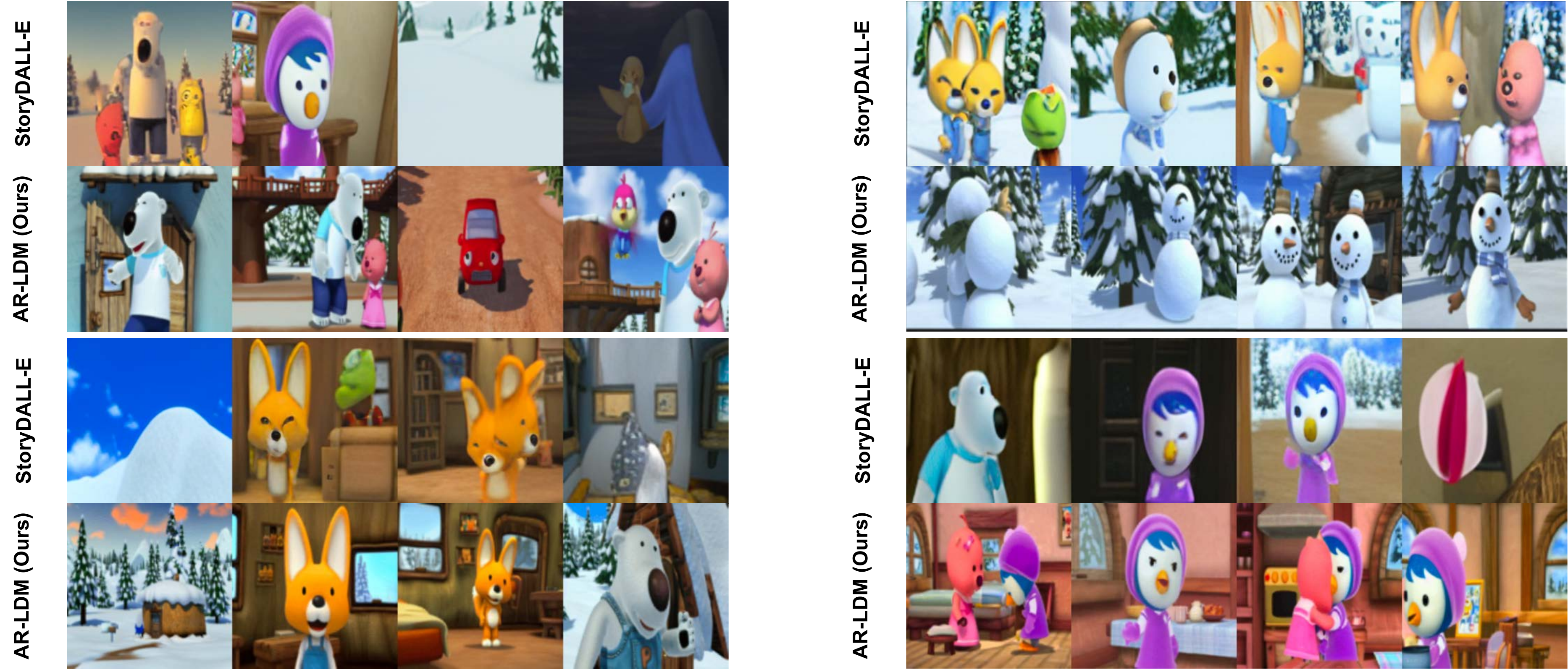}
\caption{Cases on PororoSV.}
\end{subfigure}
\vfill
\begin{subfigure}{\linewidth}
\centering
\includegraphics[height=0.3\textheight]{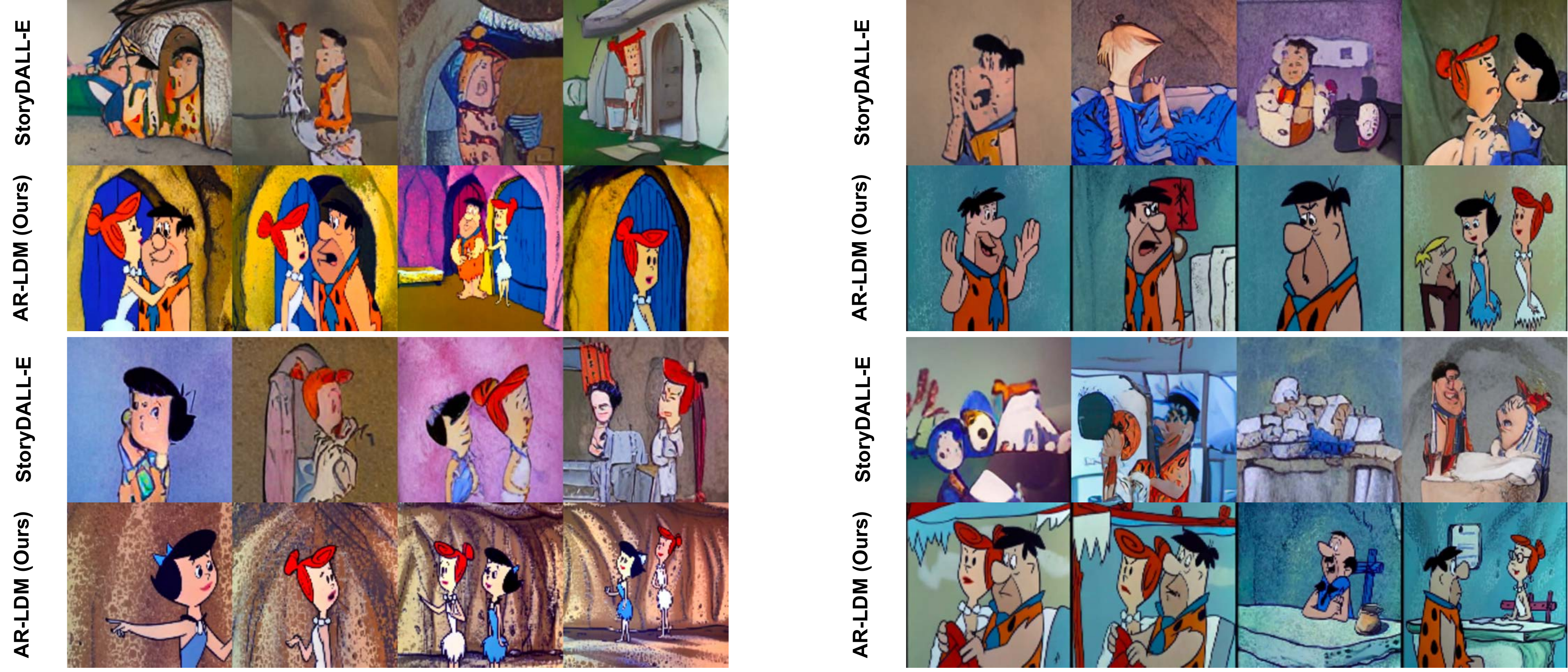}
\caption{Cases on FlintstonesSV.}
\end{subfigure}
\vfill
\begin{subfigure}{\linewidth}
\centering
\includegraphics[height=0.3\textheight]{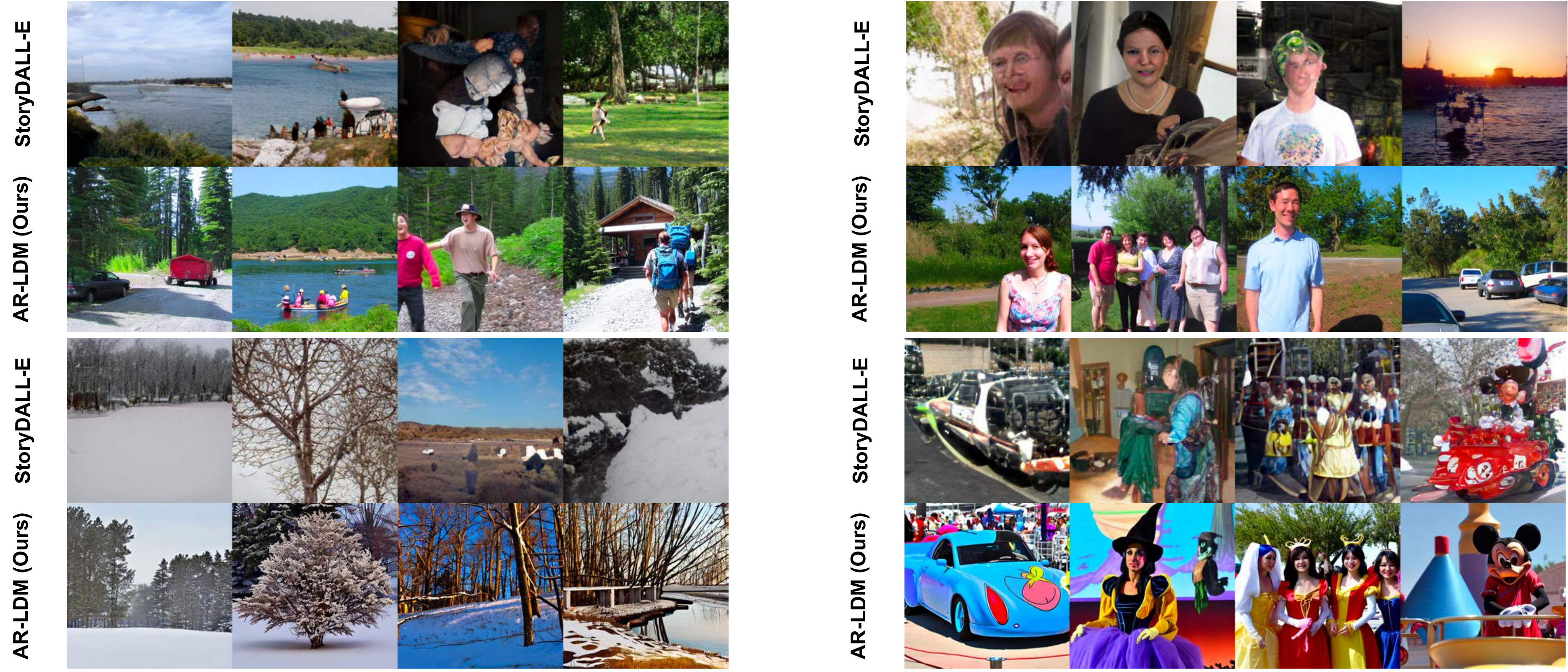}
\caption{Cases on VIST-SIS.}
\end{subfigure}
\caption{Cases that AR-LDM \textbf{wins} StoryDALL·E in human evaluation regarding \textbf{consistency}.}
\label{fig:win_consistency}
\end{figure*}

\begin{figure*}[!h]
\centering
\begin{subfigure}{\linewidth}
\centering
\includegraphics[width=0.78\linewidth]{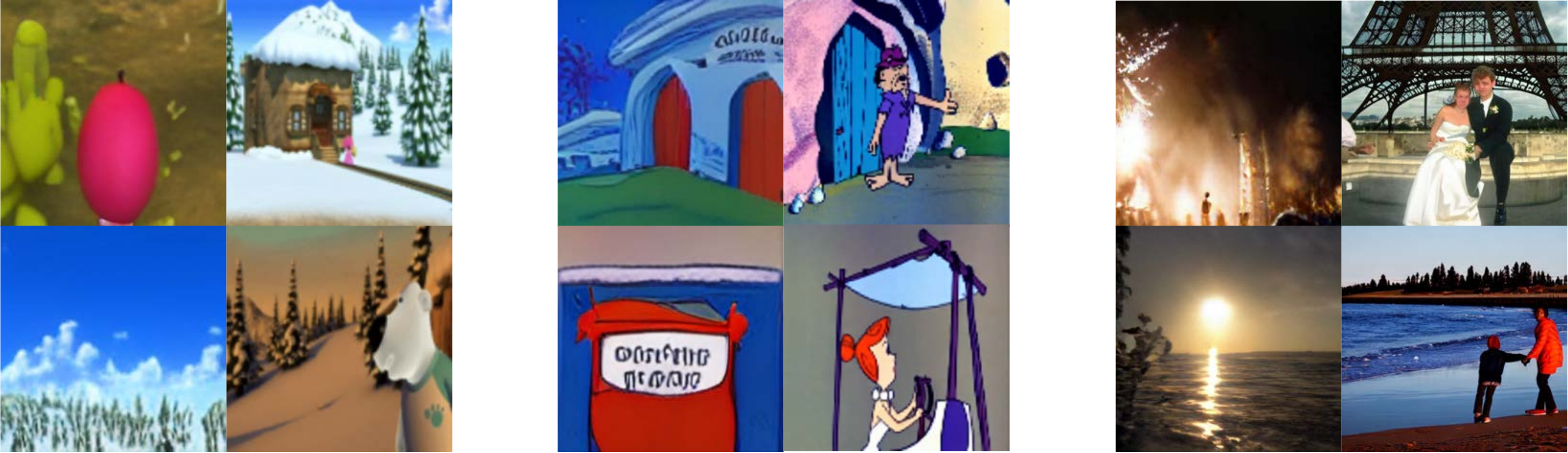}
\caption{Cases regarding visual quality. The left ones are synthesized by StoryDALL·E, and the right ones are synthesized by AR-LDM.}
\end{subfigure}
\vfill
\begin{subfigure}{\linewidth}
\centering
\includegraphics[width=0.78\linewidth]{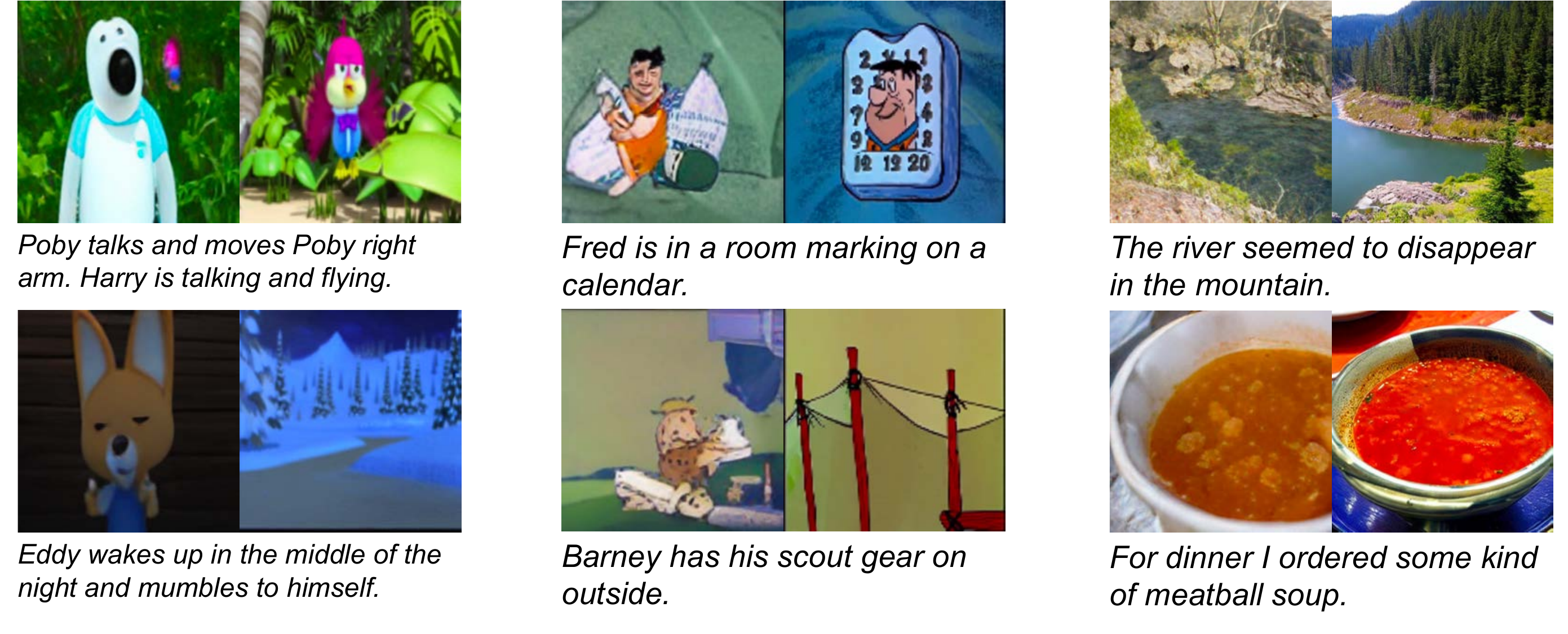}
\caption{Cases regarding relevance. The left ones are synthesized by StoryDALL·E, and the right ones are synthesized by AR-LDM.}
\end{subfigure}
\vfill
\begin{subfigure}{\linewidth}
\centering
\includegraphics[width=0.84\linewidth]{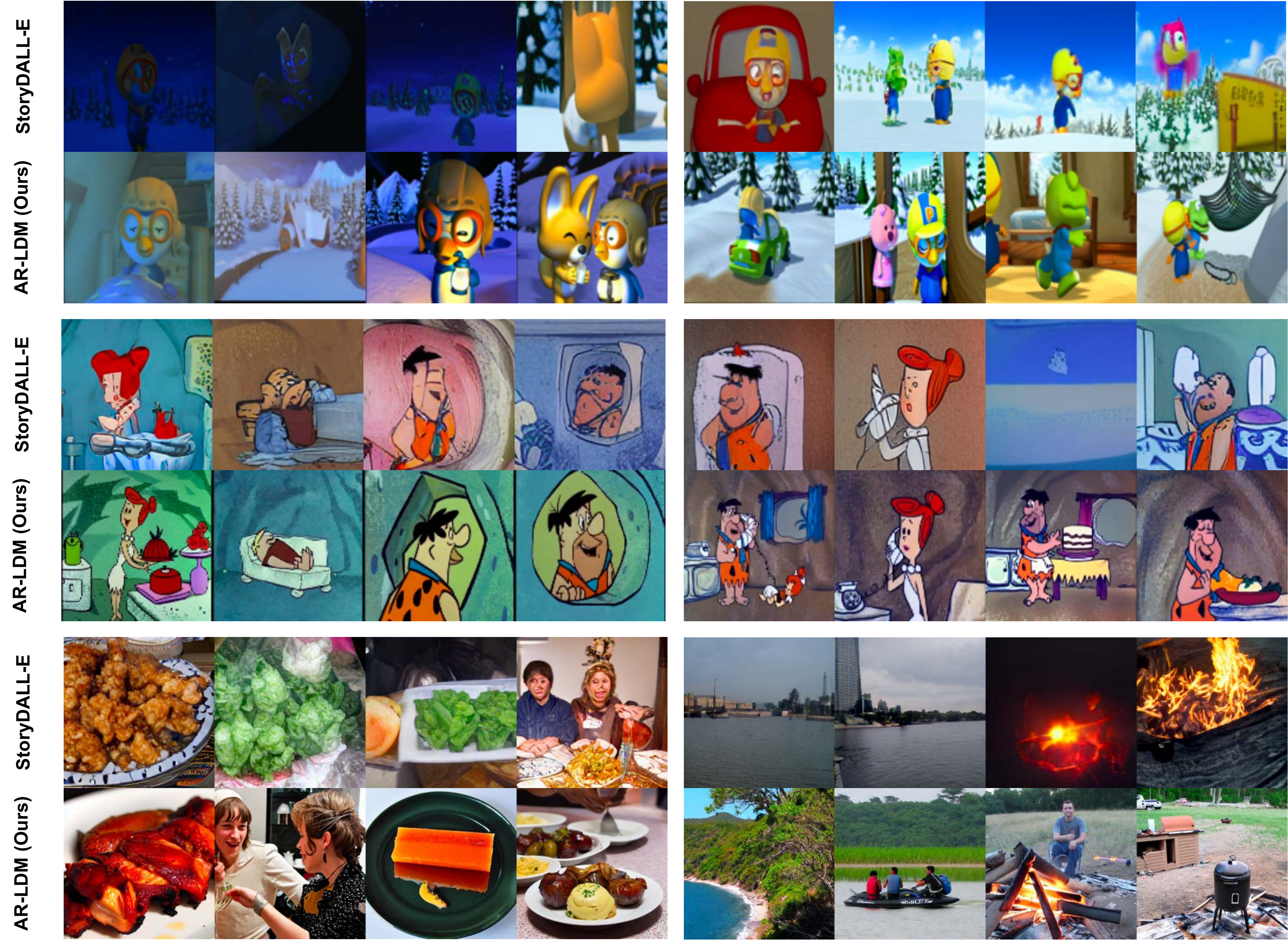}
\caption{Cases regarding consistency.}
\end{subfigure}
\caption{Cases that AR-LDM \textbf{loses} StoryDALL·E in human evaluation.}
\label{fig:lose}
\end{figure*}

In this section, we provide some cases in our human evaluation. Specifically, \cref{fig:win_quality}, \cref{fig:win_relevance}, and \cref{fig:win_consistency} show the cases that AR-LDM wins StoryDALL·E regarding visual quality, relevance, and consistency, respectively. \cref{fig:lose} show the cases that AR-LDM loses StoryDALL·E in human evaluation.

\clearpage
\section{Additional Synthesized Visual Stories}
\label{sec:additional_synthesized_visual_stories}
In this section, we provide additional synthesized visual stories.
\setlength\parindent{0pt}

\subsection{PororoSV}
\label{sec:additional_pororo}
\begin{figure*}[!th]
\centering
\includegraphics[width=\linewidth]{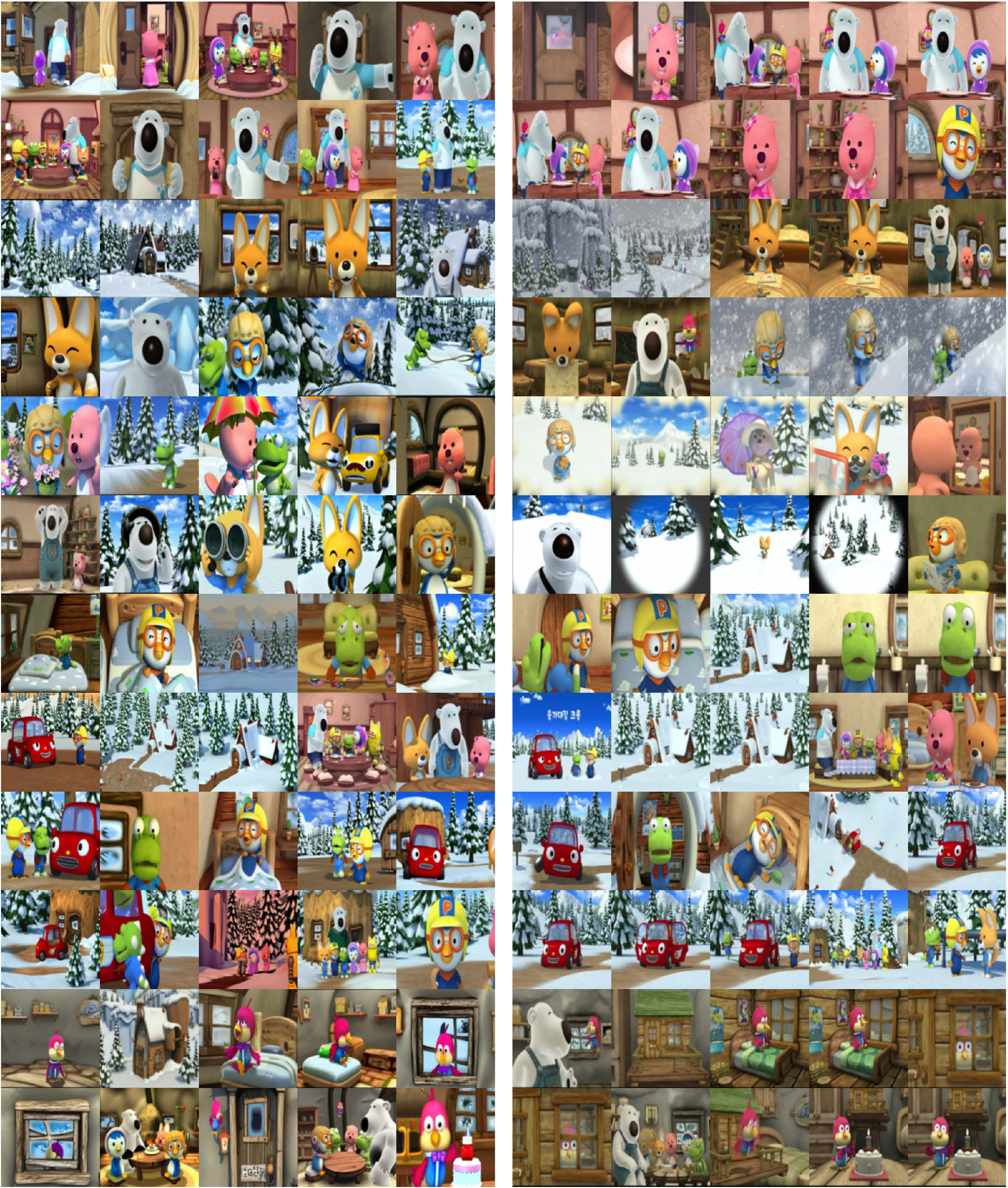}
\caption{Example of generated visual stories (left 5 frames) from AR-LDM and corresponding ground truths (right 5 frames) on PororoSV. These cases are under \textbf{story visualization} setting.}
\label{fig:additional_pororovis}
\end{figure*}

\begin{figure*}[!th]
\centering
\includegraphics[width=\linewidth]{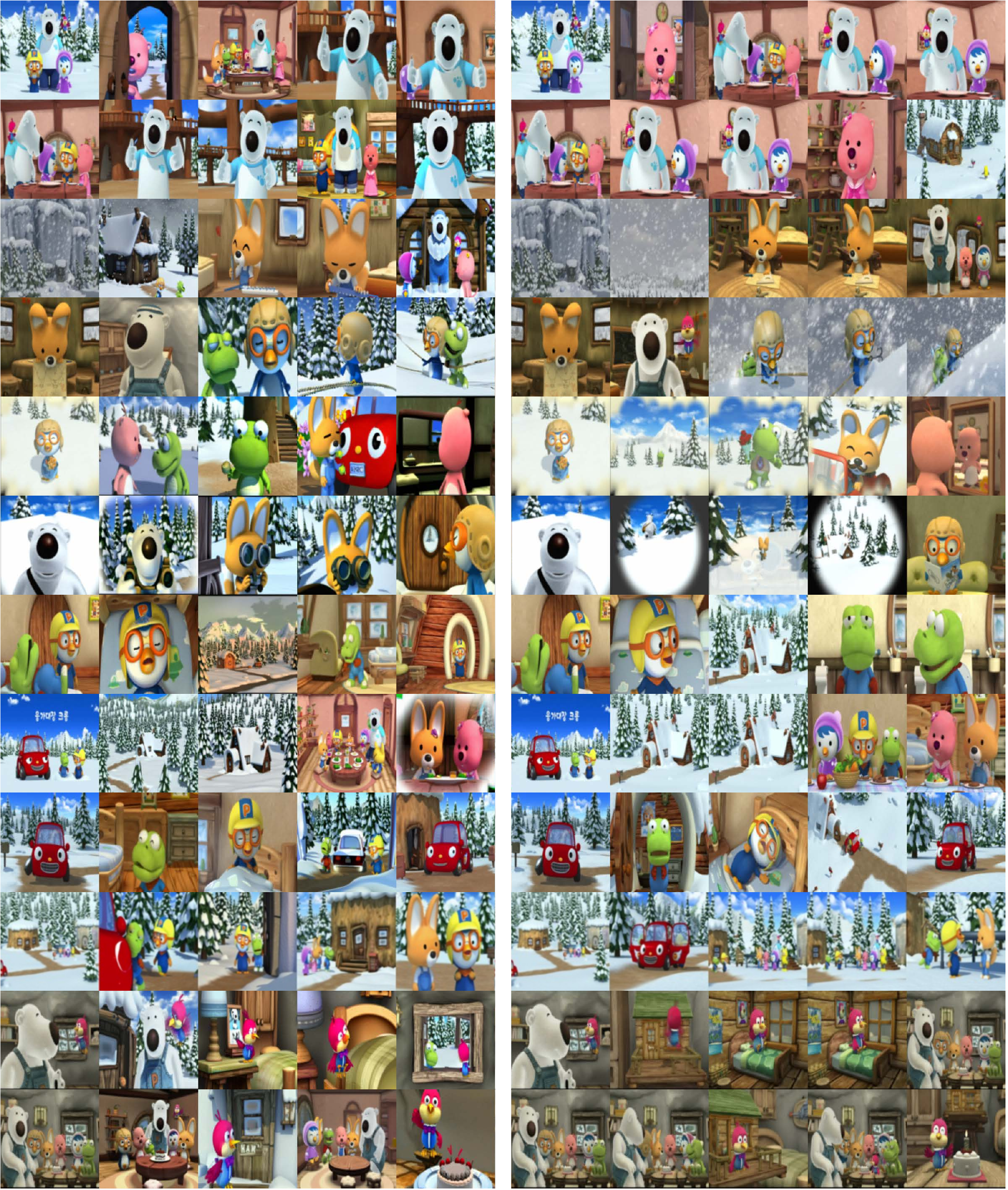}
\caption{Example of generated visual stories (left 5 frames) from AR-LDM and corresponding ground truths (right 5 frames) on PororoSV. These cases are under \textbf{story continuation} setting, which means the first frame serves as a source frame. These cases are corresponding to \cref{fig:additional_pororovis}.}
\label{fig:additional_pororo}
\end{figure*}

We provide additional cases on PororoSV in both story visualization (\cref{fig:additional_pororovis}) and story continuation (\cref{fig:additional_pororo}) settings. The corresponding captions are listed below.

\textit{\footnotesize{\textbf{Case 1:}\\
1. Petty Pororo and Poby arrives at Loopy's house.\\
2. Loopy opens door and invites Loopy friends in.\\
3. Pororo and Poby friends are finished with meal.\\
4. Poby gives the thumbs up.\\
5. Poby and Petty give the thumbs up. Loopy is happy.\\
\textbf{Case 2:}\\
1. Pororo and Poby friends are finished with meal.\\
2. Poby gives the thumbs up.\\
3. Poby and Petty give the thumbs up. Loopy is happy.\\
4. Loopy suggests to go outside.\\
5. Pororo agrees and smiles with Poby hand.\\
\textbf{Case 3:}\\
1. The weather is snowy and windy. There isn't any person in this scene.\\
2. The weather is snowy and windy and the weather is going even worse. Two characters are running toward the cabin.\\
3. Eddy is now at the room. Eddy uses pencil ruler and papers. Eddy seems satisfied with his work.\\
4. Eddy is in the room. Eddy uses pencil ruler and papers. Eddy turns his head right side.\\
5. Poby Petty Loopy and Harry are gathering in the cabin. The weather is snowy and windy. Poby thinks that emergency situation happens.\\
\textbf{Case 4:}\\
1. Eddy keep holding his picture and explains his idea to Poby. Because of the snowy weather we have no choice but to rescue Pororo and Crong by airship. Therefore Eddy resorts to Poby that they need Poby's help.\\
2. Poby seems surprised because Poby doesn't expect that Poby will be needed in this situation. After hearing from Eddy Poby turns his head to the left side.\\
3. Pororo and Crong are in the middle of the mountain. They seem tired and exhausted. Pororo close his eyes with long hard thinking.\\
4. Pororo closes his eyes with long hard thinking. The weather is snowy and it becomes worse. Pororo can't find any other solutions except rope to get out of this mountain. Pororo and Crong are stuck in this mountain so Pororo tries to use rope.\\
5. Pororo and Crong try to pull the rope to overcome this situation. However it is hard to fully apply their force.\\
\textbf{Case 5:}\\
1. In Loopy's imagination Pororo comes to her with flowers.\\
2. From far away Crong also comes to Loopy.\\
3. Loopy is wearing a fine costume and is holding a parasol. Crong gives her flowers.\\
4. Eddy with his mustache and with his car presents flowers to Loopy.\\
5. Loopy stands in front of the mirror and checks herself.\\
\textbf{Case 6:}\\
1. Poby feels ashamed and wants that nobody saw him falling down.\\
2. Seeing Poby through the telescope Eddy secretly smiles and talks to himself that Eddy saw Poby falling down.\\
3. Eddy is interested in seeing things and friends through telescope. Eddy brings telescope and goes to the mountain to observe his friends more.\\
4. Up on the mountain Eddy chooses a target. It is Pororo. Eddy looks through the telescope.\\
5. Pororo was reading a book. Loopy calls him outside his home. Pororo hears it and looks at the door if anyone came.\\
\textbf{Case 7:}\\
1. Pororo after looking Crong goes to bed says good night to Crong.\\
2. Pororo is lying on the bed.\\
3. The morning came. It got bright.\\
4. Crong tried but could not make number two.\\
5. Pororo called Crong from outside.\\
\textbf{Case 8:}\\
1. The car Pororo and Crong are on the ground.\\
2. There is Pororo's house in the forest.\\
3. Pororo's house in covered with snow.\\
4. Pororo and friends are eating lunch.\\
5. Eddy and Loopy are sitting next to each other.\\
\textbf{Case 9:}\\
1. The car tells Crong that Pororo and Crong are to meet at Eddy's house for a picnic.\\
2. Remembering the appointment Crong was surprised.\\
3. While Pororo is sleeping Crong calls Pororo.\\
4. Pororo and Crong came out of the house. Pororo and Crong ride in the car.\\
5. The car arrives at Eddy's house.\\
\textbf{Case 10:}\\
1. The car arrives at Eddy's house.\\
2. Pororo and Crong are coming out of the car.\\
3. Pororo and Crong greet friends friends.\\
4. All the friends are standing in front of Eddy's house.\\
5. Pororo is asking about something.\\
\textbf{Case 11:}\\
1. Poby and Harry face each other with smile. Harry looks excited.\\
2. Harry's house lays down on one side of the Poby's house.\\
3. Harry sits down on the bed. Harry really skips about for joy.\\
4. Harry sits down on her bed with joy.\\
5. Harry is looking out of the window.\\
\textbf{Case 12:}\\
1. Harry is looking out of the window.\\
2. Pororo and his friends sit round with joy. Cake lies on the table.\\
3. Harry stand up in front of Harry's house.\\
4. Pororo and his friends are sitting around the table. They congratulates Harry's new house.\\
5. Harry stands up beside the cake. Harry is really happy.\\
}}

\clearpage

\subsection{FlintstonesSV}
\begin{figure*}[!th]
\centering
\includegraphics[width=\linewidth]{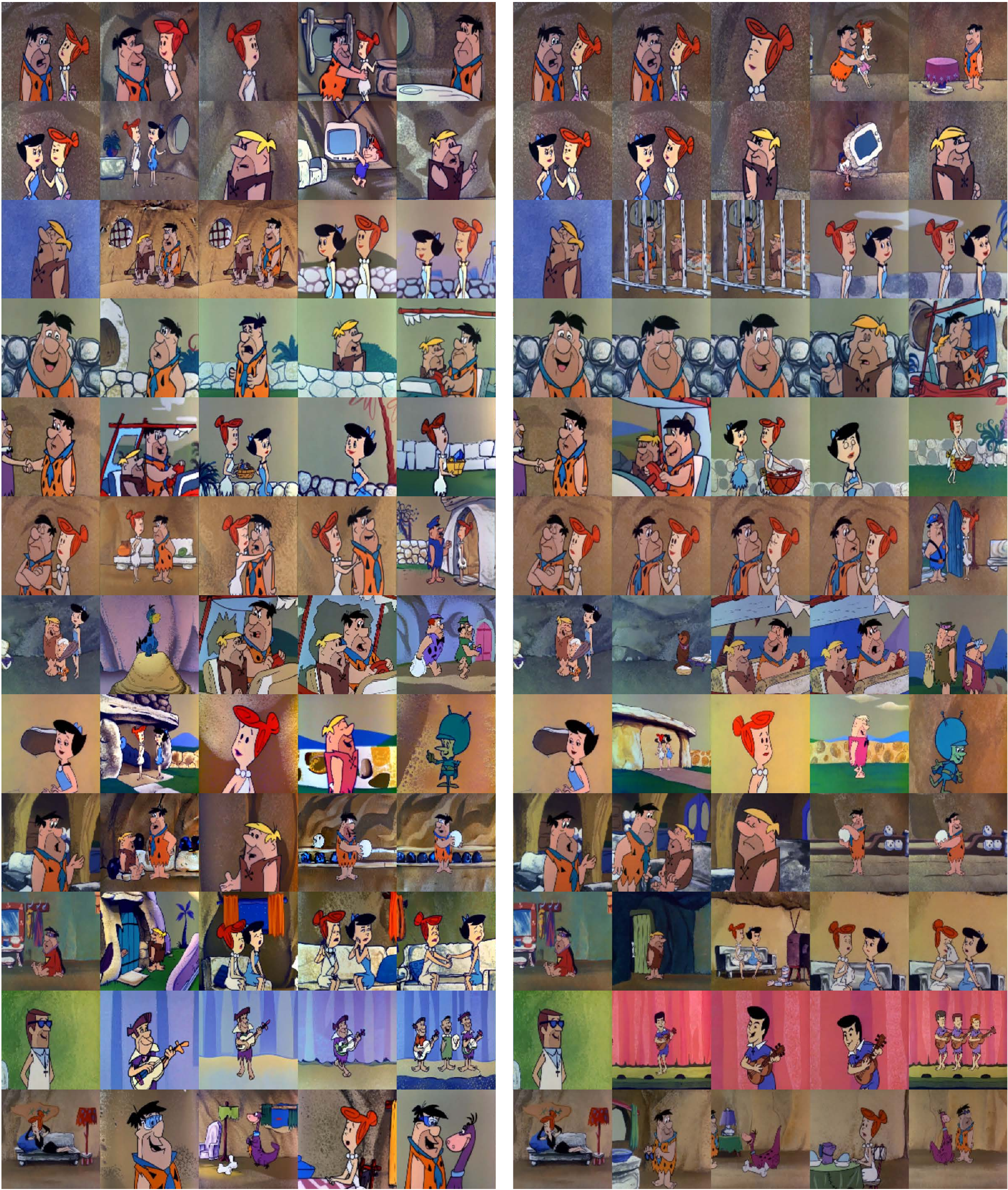}
\caption{Example of generated visual stories (left 5 frames) from AR-LDM and corresponding ground truths (right 5 frames) on FlintstonesSV. These cases are under \textbf{story continuation} setting, which means the first frame serves as a source frame.}
\label{fig:additional_flintstones}
\end{figure*}

As shown in \cref{fig:additional_flintstones}, We provide additional cases on FlintstonesSV in the story continuation setting. The corresponding captions are listed below.

\textit{\footnotesize{\textbf{Case 1:}\\
1. Fred and Wilma are standing in a room. Wilma speaks to Fred.\\
2. Fred and Wilma are in the room, Fred is talking. Wilma reaches to hug him.\\
3. Wilma is speaking in the room.\\
4. Wilma and Fred are in a room. Fred is grabbing Wilma by the hips and pushing her through the room.\\
5. Fred is standing in the dining room, waiting to be served.\\
\textbf{Case 2:}\\
1. Wilma and Betty are standing in a room speaking.\\
2. Wilma and Betty are standing in a room talking.\\
3. Barney turns to talk to someone behind him in the room with an angry look on his face.\\
4. A small boy holding the stone is in the living room. He holds the tv as he walks.\\
5. Barney is in the living room talking sternly and wagging his finger.\\
\textbf{Case 3:}\\
1. Barney is standing in a room. He speaks and looks tired.\\
2. Fred and Barney are in jail. Fred is explaining something to Barney while the two of them are standing in a cell behind bars.\\
3. Fred and Barney are in jail. Fred opens his arms and speaks. Then Barney responds.\\
4. Wilma and Betty are walking through a yard together.\\
5. Wilma and Betty are happily walking next to each other outside. Betty is talking while Wilma is listening.\\
\textbf{Case 4:}\\
1. Fred is outside. He talks with a tear in his eye.\\
2. Fred is in the backyard in front of the stone wall talking to someone.\\
3. Fred is standing outside while crying and talking.\\
4. Barney is in the yard. He is talking and gesturing with his hand.\\
5. Fred and Barney are sitting in a car talking.\\
\textbf{Case 5:}\\
1. Fred is awkwardly shaking a man's hand in the room.\\
2. Fred is driving a car down the road. He is speaking to Barney.\\
3. Wilma is holding a basket out in the yard. She is listening to Betty. They are standing in front of a stone fence.\\
4. Betty is standing in her yard, talking to someone off camera right.\\
5. Wilma is walking through the yard. She is carrying a basket. While walking she is speaking to someone behind her.\\
\textbf{Case 6:}\\
1. Wilma talking to Fred in a room.\\
2. Wilma talks to Fred with her hands on his back in the living room.\\
3. Wilma and Fred are in the room. Fred looks upset and says something. Wilma holds his shoulders and says something.\\
4. Fred and Wilma are standing in a room. Fred speaks while Wilma holds onto his shoulder.\\
5. A police officer in uniform with a long skinny nose is standing in a doorway talking to Wilma.\\
\textbf{Case 7:}\\
1. Barney laughs and talks to bamm bamm while they walk with Betty outside.\\
2. the animal is standing on a rock and clapping its fins in a cave.\\
3. Fred and Barney are in the car talking to each other.\\
4. Fred is driving a car while Barney rides in the passenger seat. They talk briefly.\\
5. The man in green is outside holding a bag and wearing a hat. He pushes the man with glasses in pink clothes and purple tie.\\
\textbf{Case 8:}\\
1. Betty is standing in a room. She speaks, leans back, and then begins to race off.\\
2. Wilma and Betty are standing on the driveway and looking inside the garage.\\
3. Wilma in the room talking to someone.\\
4. Barney walks through the yard in a pink shirt. He looks back while talking, then looks forward and laughs with his eyes closed.\\
5. The great gazoo floats in the room as he speaks.\\
\textbf{Case 9:}\\
1. Fred is in the living room. He is talking to someone.\\
2. Barney is sitting on a bench in a bowling alley while Fred stands and they have a conversation.\\
3. Barney is in a room talking.\\
4. first, Fred looks down at the bowling ball at the bowling alley with a funny confused look and sticking his tongue out. Then, Fred looks behind him while still holding the bowling bowl.\\
5. Fred is in the bowling alley. He stands with a while bowling ball in his hands and then runs to the left in preparation to bowl. He speaks to someone off screen to the right.\\
\textbf{Case 10:}\\
1. Fred is sitting on the ground in the dressing room. He is wearing a purple eye mask and red suit with a purple cape.\\
2. Barney is talking near the doorway.\\
3. Betty and Wilma are sitting in a living room. Wilma begins to cry and brings a tissue to her face. Then Betty turns to look at Wilma.\\
4. Wilma and Betty are sitting on a couch in the living room. Wilma speaks to Betty and cries into a handkerchief.\\
5. Wilma and Betty are sitting on a couch in the living room. Wilma is crying and wiping her tears with a handkerchief while Betty speaks to her.\\
\textbf{Case 11:}\\
1. The fancy man in white suit is in the room, pointing to the ceiling as he talks.\\
2. The musician with guitar is on stage. He is singing.\\
3. There is a man playing guitar on a stage.\\
4. The man with blue short playing guitar is on the stage. He is dancing.\\
5. Three men are on stage playing guitar and singing. There is a man with black hair, a man with brown hair, and a man with red hair.\\
\textbf{Case 12:}\\
1. Wilma yells at Fred while lying on the couch in the living room.\\
2. Fred is in the room and puts on sunglasses.\\
3. Dino is in the living room. He is wagging his tail while laying on a bone when Fred walks by.\\
4. Wilma sits at the table in a room. She is talking.\\
5. Fred is wearing blue glasses standing in an empty room talking to Dino.
}}

\clearpage

\subsection{VIST-SIS}

\begin{figure*}[!th]
\centering
\includegraphics[width=\linewidth]{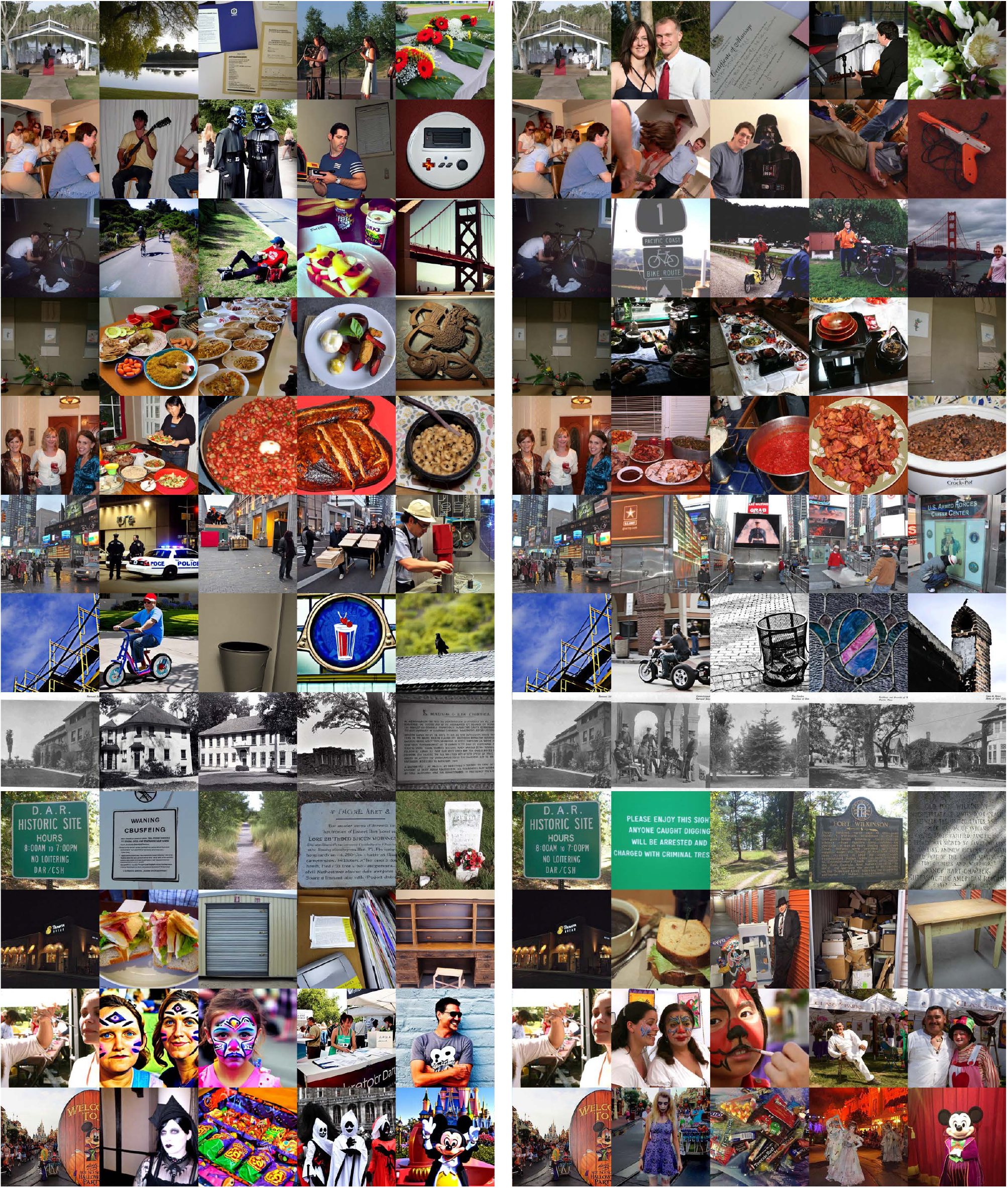}
\caption{Example of generated visual stories (left 5 frames) from AR-LDM and corresponding ground truths (right 5 frames) on VIST-SIS. These cases are under \textbf{story continuation} setting, which means the first frame serves as a source frame.}
\label{fig:additional_vistsis}
\end{figure*}

As shown in \cref{fig:additional_vistsis}, We provide additional cases on VIST-SIS in the story continuation setting. The corresponding captions are listed below.

\textit{\footnotesize{\textbf{Case 1:}\\
1. I went to the wedding last week.\\
2. It was on the lake.\\
3. I brought all the necessary paperwork.\\
4. The live band was very good.\\
5. I bought many flowers for the couple.\\
\textbf{Case 2:}\\
1. On the night of the party everyone was so excited to see each other.\\
2. A few of the guys broke out the guitar and started to play some tunes.\\
3. My friend james got a photo standing next to his favorite character darth vader.\\
4. For most of the night we decided to play retro video games.\\
5. The gun I used while playing the nintendo game dunk hunt.\\
\textbf{Case 3:}\\
1. Fixing up his bike.\\
2. Following the pacific coast bike route.\\
3. Taking a lunch break.\\
4. Eating fruits and sandwiches.\\
5. About to cross the san francisco bridge.\\
\textbf{Case 4:}\\
1. From the entrance, the decorations told us we were having a traditional japanese meal.\\
2. The low tables had all sorts of delicious food our hosts had prepared.\\
3. But that was just the beginning, we saw as we walked towards a larger table with a fest spread upon it.\\
4. Soon we'd consumed everything -- it was all so tasty, and interesting, too.\\
5. We took a moment before we left to take a closer look at the art on the walls, and found the meaning for our hosts; I guess it's rude to ask to come back right away!\\
\textbf{Case 5:}\\
1. The three sisters gathered to celebrate thanksgiving.\\
2. Their mom made a lot of good food for dinner.\\
3. She even made tons of mashed potatoes and spaghetti sauce.\\
4. The plate of pulled pork is always a crowd pleaser.\\
5. Black eyed peas are delicious as well.\\
\textbf{Case 6:}\\
1. The city is very crowded with people.\\
2. The cops watch over to keep it safe.\\
3. We are getting ready to set up the party.\\
4. We are carrying the table.\\
5. Ken is adjusting everything just right.\\
\textbf{Case 7:}\\
1. A bright blue sky to start off the day.\\
2. Taking the trike on a drive across town.\\
3. Passed an odd shaped trash can that could have seen better days.\\
4. Saw a beautiful glass sign that was perfect for the old lady but the owner refused to sell.\\
5. Above a crow watches from the chimney.\\
\textbf{Case 8:}\\
1. Today the class was looking at history.\\
2. We saw a lot of old looking pictures.\\
3. Some of the pictures were of familiar places where we lived.\\
4. However a lot of these places were of old parks that had been torn down.\\
5. History sure is interesting. there's a lot to learn!\\
\textbf{Case 9:}\\
1. We visited some historic sites on our trip.\\
2. They had a lot of warning signs when we arrived.\\
3. After that we traveled the paths at the site.\\
4. Then we came across signs that told us the history of the site.\\
5. After that we came across some old graves at the site.\\
\textbf{Case 10:}\\
1. Before going to my storage locker I stopped off for some food.\\
2. This place always serves up great sandwiches.\\
3. After eating I made it the storage unit.\\
4. It is so full of papers and other things that I do not need.\\
5. I found this old desk in there that I forgot I had.\\
\textbf{Case 11:}\\
1. I love going to the art fest.\\
2. Me and my mom got our faces painted.\\
3. My daughter was transformed with the face paint.\\
4. So many people had booths there.\\
5. My husband even had a good time.\\
\textbf{Case 12:}\\
1. Greetings from the disneyland Halloween party!\\
2. I dressed like a goth.\\
3. Surprisingly, I got a fair amount of candy!\\
4. I also saw so spooky ghost dancers,\\
5. And even count mickey! hope to see you thanksgiving!\\
}}

\clearpage

\subsection{VIST-DII}

\begin{figure*}[!th]
\centering
\includegraphics[width=\linewidth]{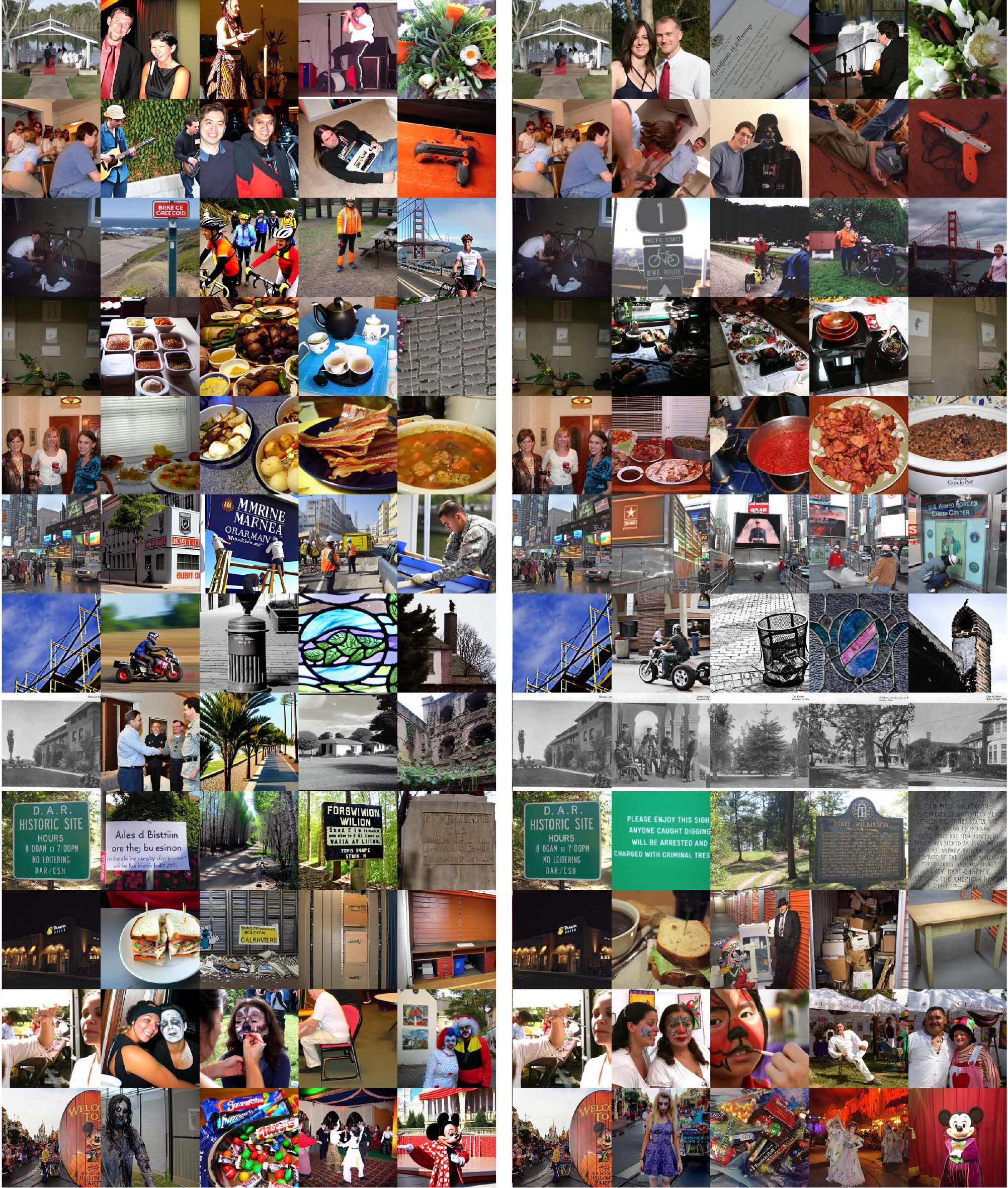}
\caption{Example of generated visual stories (left 5 frames) from AR-LDM and corresponding ground truths (right 5 frames) on VIST-DII. These cases are under \textbf{story continuation} setting, which means the first frame serves as a source frame. These cases are corresponding to \cref{fig:additional_vistsis}}
\label{fig:additional_vistdii}
\end{figure*}

As shown in \cref{fig:additional_vistdii}, We provide additional cases on VIST-DII in the story continuation setting. These cases are corresponding to \cref{fig:additional_vistsis}, the captions are listed below.

\textit{\footnotesize{\textbf{Case 1:}\\
1. A man walking on a red carpet in an outdoor building set up for a wedding.\\
2. A man and a woman pose together, she in a black dress, he in a red tie.\\
3. The lady that performed the wedding ceremony was a native american.\\
4. An entertainer performing music and singing at an event.\\
5. A bouquet of flowers features white flowers and succulent greens.\\
\textbf{Case 2:}\\
1. Group photo with a group of females wearing white tops and sunglasses.\\
2. One man is playing the guitar while the other watches on.\\
3. Very happy young man getting his picture taken with darth vader.\\
4. Long haired man lying on the floor playing a videogame.\\
5. A nintendo gun-shaped video game controller lays on a burnt orange carpet.\\
\textbf{Case 3:}\\
1. A young man is working on fixing his pedal bike.\\
2. A sign states that a bike route to the pacific coast of california is up ahead.\\
3. To people wearing helmets standing with their bicycles.\\
4. A person wearing an orange jacket and helmet stands next to a bicycle and picnic table.\\
5. A woman bicyclist posing in front of the san francisco bay bridge.\\
\textbf{Case 4:}\\
1. This space has three pictures hanging on the wall and a plant in a vase and two small sculptures on a table.\\
2. Square table with food on top in bowls and containers.\\
3. A very big meal all laid out on a long table.\\
4. A tray holding dishes and a tea kettle sits on a table set with food.\\
5. A wall has scrolls hanging down it and flowers are next to it.\\
\textbf{Case 5:}\\
1. Three woman holding wine in the foyer of a nice house.\\
2. Plates of food sitting on a kitchen table in front of window with white blinds.\\
3. Two bowls that are filled with different types of foods.\\
4. A plate of piled cooked crispy bacon ready to be eaten.\\
5. A crock pot filled with delicious looking warm soup.\\
\textbf{Case 6:}\\
1. A mob of people wait for the light to change to walk across a city street.\\
2. Small silver building on a busy street with one guard and a us army advertisement.\\
3. Workmen work at something below a marine advertisement.\\
4. Construction workers making repairs on the side of the road in a big city.\\
5. A worker repairing a window frame of the u.s. armed forces career center.\\
\textbf{Case 7:}\\
1. The bars of the structure are yellow and the sky is extremely clear.\\
2. A man in helmet rides a motorcycle with three wheels.\\
3. A black and white image of a metal grated trash can sitting on a brick sidewalk.\\
4. A section of a stained glass window displaying a blue oval with a white stripe detailed with plant like shapes.\\
5. An old building with a brick chimney in marginal shape, with a crow perched on top.\\
\textbf{Case 8:}\\
1. A scan of a black and white photo of a house is labeled ``Harvard School''.\\
2. The military personnel meeting was held at the home of the highest ranking official.\\
3. Different kinds of trees line the walkway including a palm tree.\\
4. A black and white picture of a yard with trees and a house in the background.\\
5. Vines partially cover the walls of a large, old building.\\
\textbf{Case 9:}\\
1. A green sign with white lettering describing the hours and rules for the D.A.R. historic site.\\
2. The sign was displayed on a poster so all could see.\\
3. A dirt road runs through the forest of green trees.\\
4. The fort wilkinson sign located in the state of georgia.\\
5. A carving in stone gives remembrance to the citizens during the american revolution.\\
\textbf{Case 10:}\\
1. The outside of a panera bread restuarant is lit up at night.\\
2. A club sandwich sits beside a bowl of soup in this tempting meal.\\
3. A pile of items, including cartoon character cutouts and signs, is placed in a row of storage lockers.\\
4. A storage unit filled numerous boxes filled with various items.\\
5. A wooden desk sitting in the hall way of self storage.\\
\textbf{Case 11:}\\
1. A woman with blonde hair is giving another woman face paint at a fairground.\\
2. One woman leans into the ear of another while they are wearing face paint.\\
3. A girl with a dark complexion is getting her face painted for a special event.\\
4. A man is leaning to the side while sitting in a folding chair.\\
5. A man and a woman dressed as a clown pose in front of artworks displayed beneath canopies.\\
\textbf{Case 12:}\\
1. People walk through a festival in the street while a sign advertises a mickey mouse Halloween party.\\
2. The person is dressed up like a zombie on the walking dead.\\
3. Random fun size candy, skittles, tootsie roll, milky way, 3 musketeers, lemon heads, m\&m's.\\
4. People in costumes are dancing on a ballroom floor.\\
5. A person in a mickey mouse costume stands by a red curtain.\\
}}

\setlength\parindent{1pc}

\clearpage
\section{Additional Unseen Character Adaptation Results}
\label{sec:additional_unseen_character_adaptation_results}
\begin{figure}[!th]
\centering
\includegraphics[width=\linewidth]{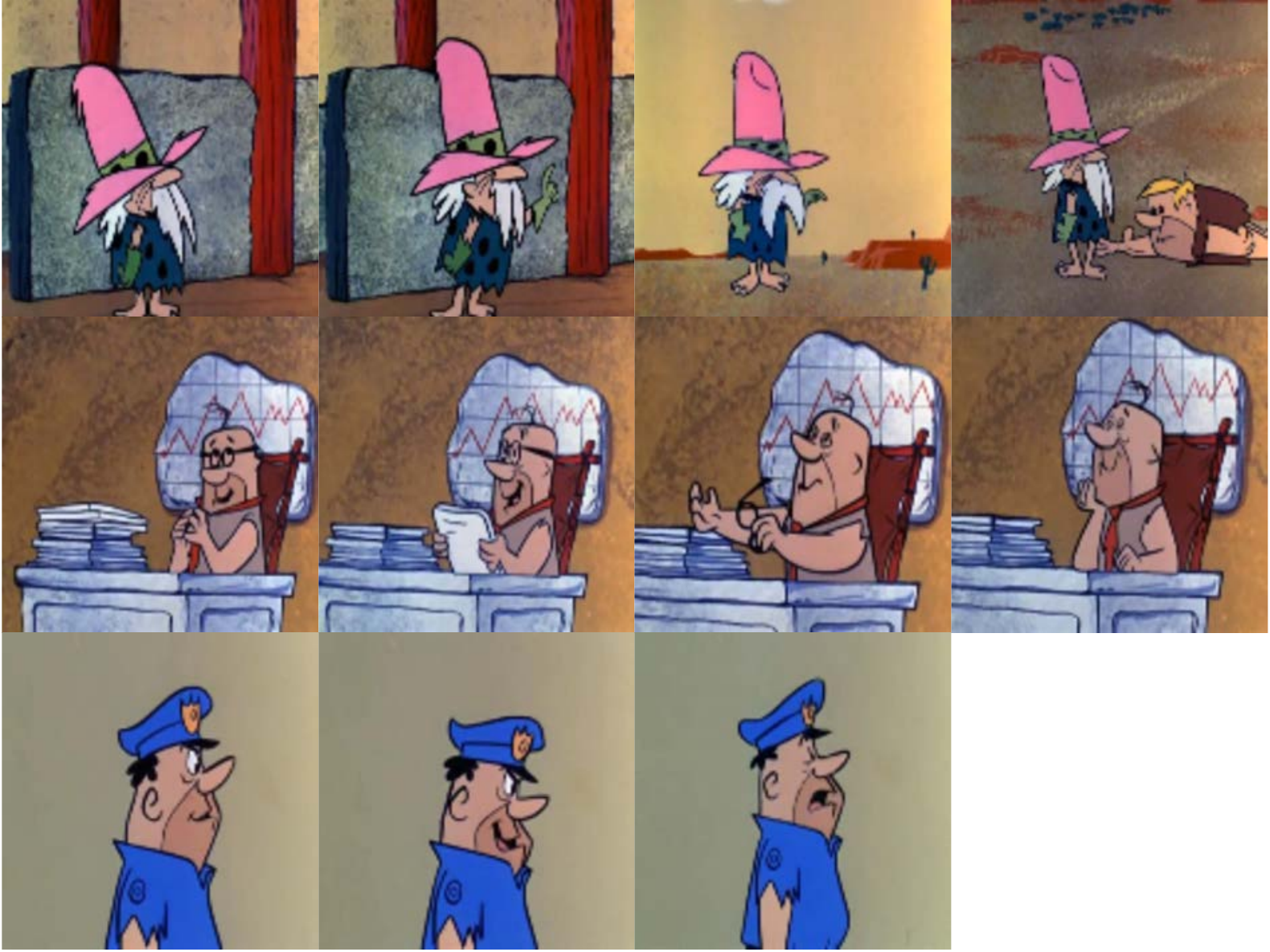}
\caption{The training images used in the adaptation cases in \cref{fig:textualinversion}, \cref{fig:unseen_cases_1}, and \cref{fig:unseen_cases_2}.}
\label{fig:unseen_cases_source}
\end{figure}

\begin{figure}[!th]
\centering
\includegraphics[width=\linewidth]{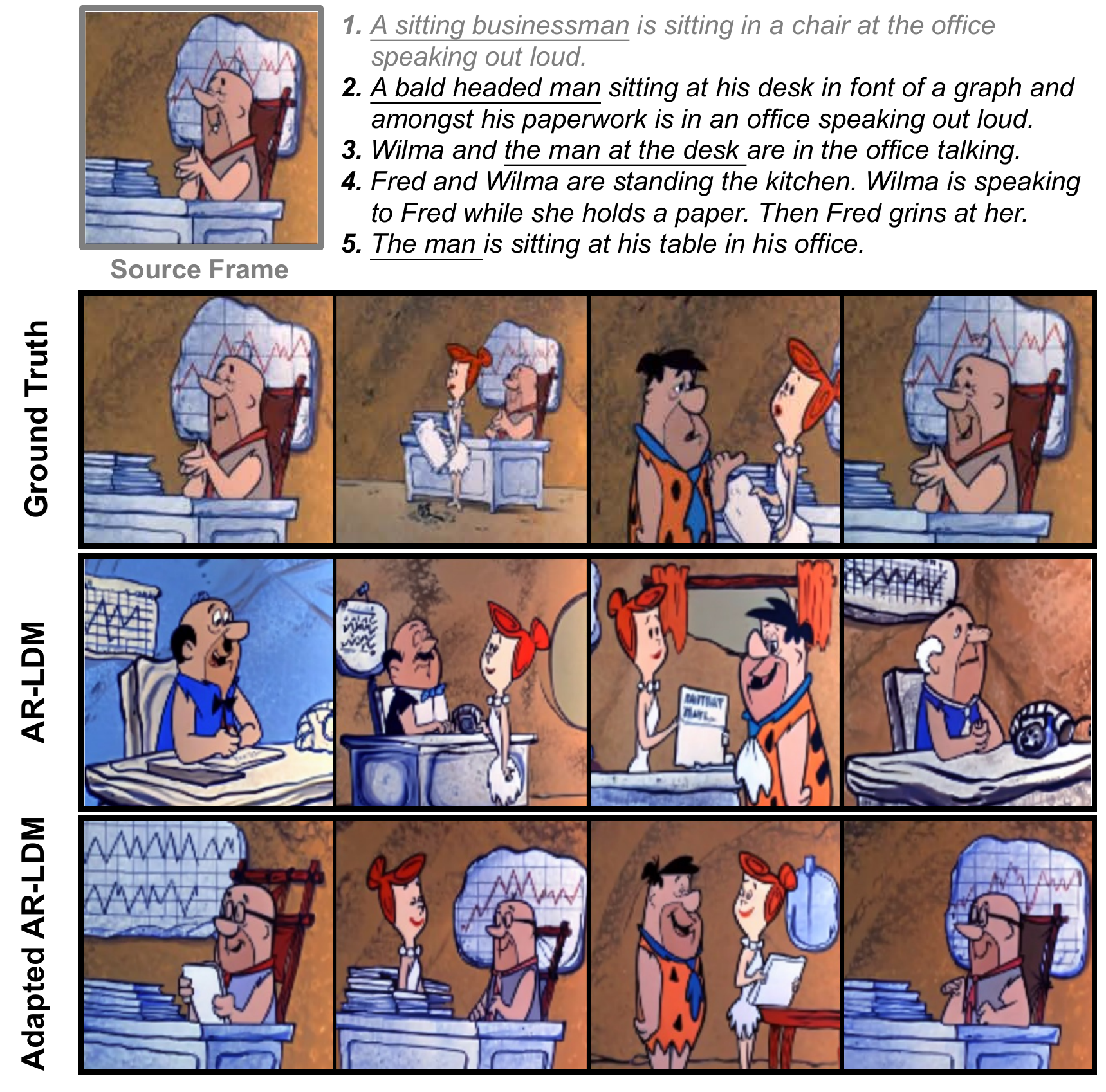}
\caption{Adaptation results for a case AR-LDM failed to properly generate on FlintstonesSV. The underlined texts refer to one specific person and can be replaced by \texttt{<char>} in adapted AR-LDM.}
\label{fig:unseen_cases_1}
\end{figure}

\begin{figure}[!th]
\centering
\includegraphics[width=\linewidth]{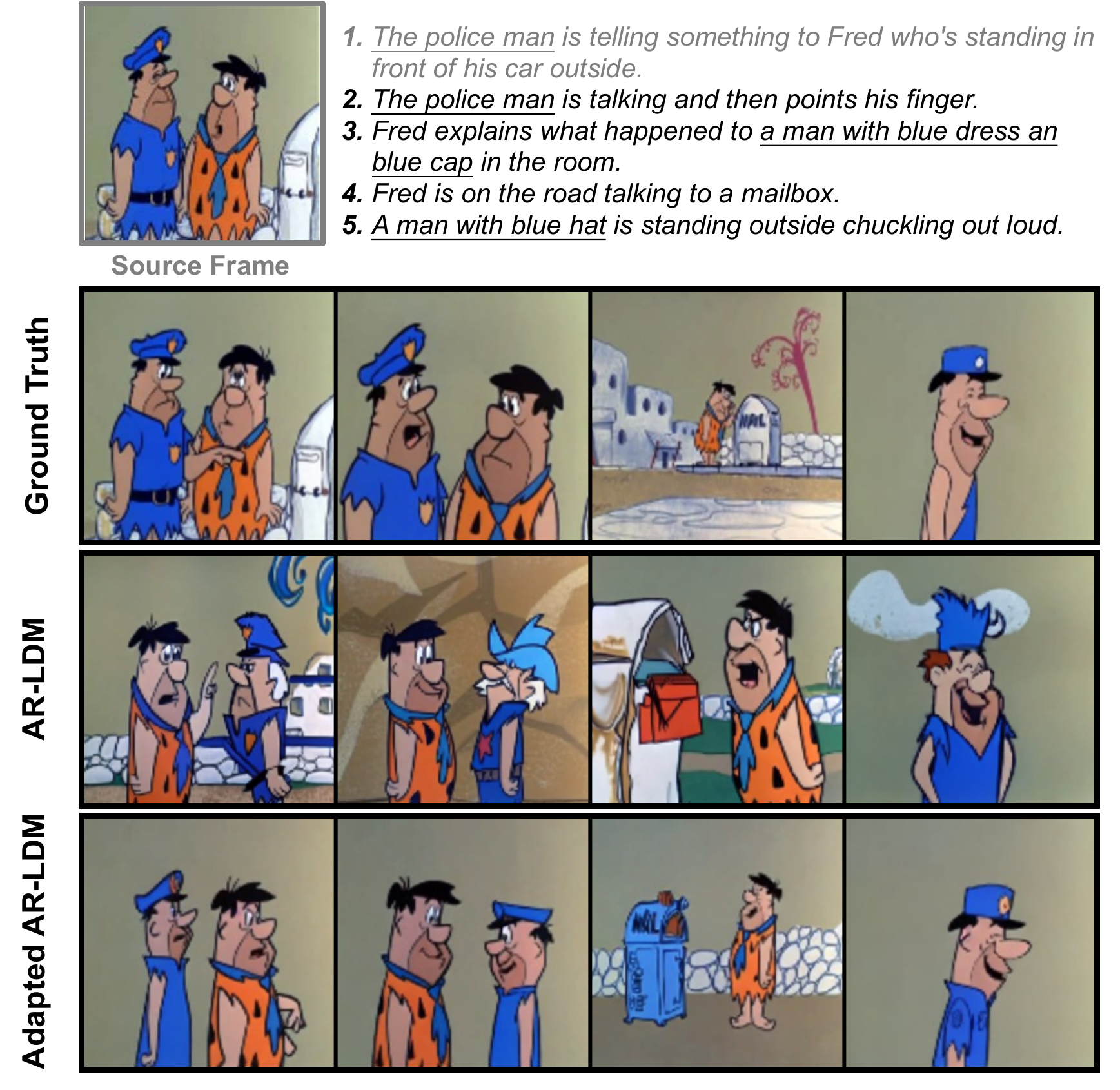}
\caption{Adaptation results for a case AR-LDM failed to properly generate on FlintstonesSV. The underlined texts refer to one specific person and can be replaced by \texttt{<char>} in adapted AR-LDM.}
\label{fig:unseen_cases_2}
\end{figure}

In this section, we provide additional cases for adaptive AR-LDM. As shown in \cref{fig:unseen_cases_1} and \cref{fig:unseen_cases_2}, AR-LDM can successfully adapt to the new character given only 3-5 images. In \cref{fig:unseen_cases_source}, We also present the training images and captions we used in the adaptation cases in \cref{fig:textualinversion}, \cref{fig:unseen_cases_1}, and \cref{fig:unseen_cases_2}, respectively.

\setlength\parindent{0pt}
\footnotesize{\textit{
\textbf{Case in \cref{fig:textualinversion}:}\\
1. \texttt{<char>} stands in the room speaking.\\
2. \texttt{<char>} is in the room. He talks as he points with his finger.\\
3. \texttt{<char>} is outside.\\
4. \texttt{<char>} is outside talking down to Barney who s laying on the ground with one of his hands stretched out above his head.\\
\textbf{Case in \cref{fig:unseen_cases_1}:}\\
1. \texttt{<char>} is sitting at his desk in the office and talking.\\
2. \texttt{<char>} speaks to himself in a office as he reads a piece of paper.\\
3. \texttt{<char>} is looking at the papers in his office speaking.\\
4. \texttt{<char>} ponders something at his office desk.\\
\textbf{Case in \cref{fig:unseen_cases_2}:}\\
1. \texttt{<char>} is outside talking.\\
2. \texttt{<char>} is outside. He speaks quickly.\\
3. \texttt{<char>} is talking outside.\\
}}
\setlength\parindent{1pc}


\clearpage
\section{Discussion}
\label{sec:additional_discussion}
We find that many failure cases can be attributed to failing to ground the entity correctly. As shown in \cref{fig:dicussion} (the captions are also given below), though AR-LDM is able to preserve the consistency of the bed across the second and the third frames, it also draws some entities in the wrong color. In the third frame, the blanket incorrectly grounds the green dress Wilma wears in the second frame. These failure groundings cause inconsistency across frames. We assume this can be attributed to BLIP's lack of the masked image modeling pre-training task. We suggest using multimodal PTMs with this pre-training task like BEiT-3~\cite{beit3} after their weight is available.

Another limitation of AR-LDM is that we can only synthesize the images one by one, while in the case below, we can observe that only in the last frame are we given enough information to generate Fred's cloth correctly. This can be alleviated by performing info-sharing between captions. For instance, we can input the whole 5 captions into the CLIP text encoder, and use the corresponding outputs to guide the diffusion model. But in practice, we find it results in confused synthesized contents, which means the self-attention will mix up the semantic meaning of different sentences. This can be improved by replacing the text encoder of AR-LDM in the future.

\begin{figure}[!ht]
\centering
\begin{subfigure}{\linewidth}
\includegraphics[width=\linewidth]{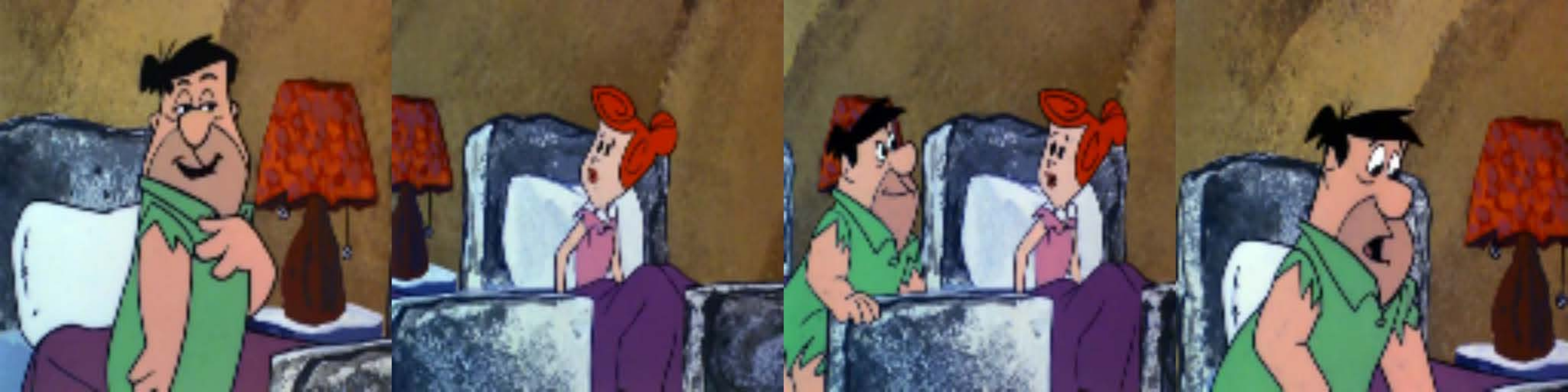}
\caption{Ground truth visual story.}
\label{fig:dicussion_gt}
\end{subfigure}
\begin{subfigure}{\linewidth}
\includegraphics[width=\linewidth]{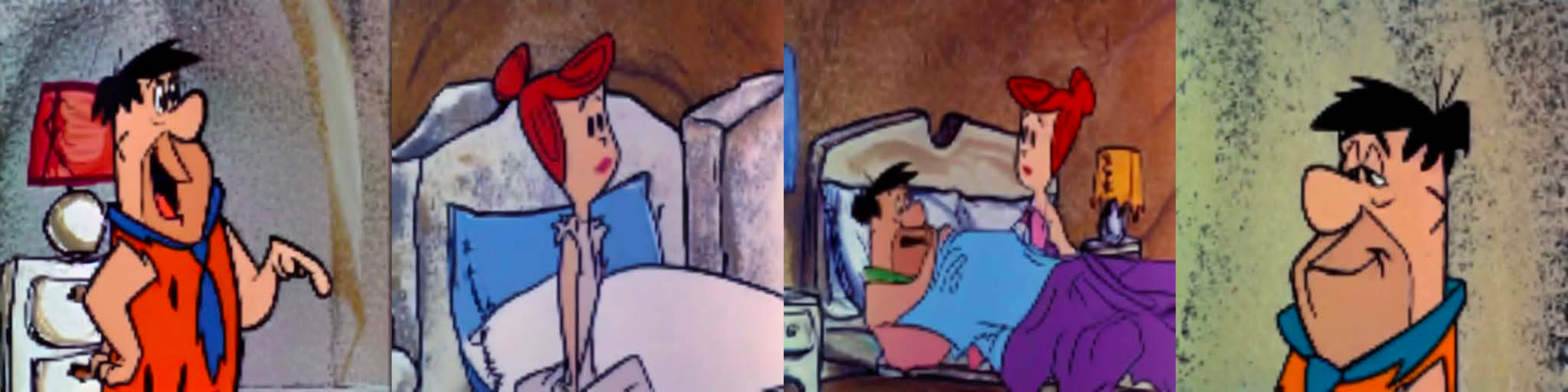}
\caption{Visual story synthesized by finetuned Stable Diffusion.}
\label{fig:dicussion_baseline}
\end{subfigure}
\begin{subfigure}{\linewidth}
\includegraphics[width=\linewidth]{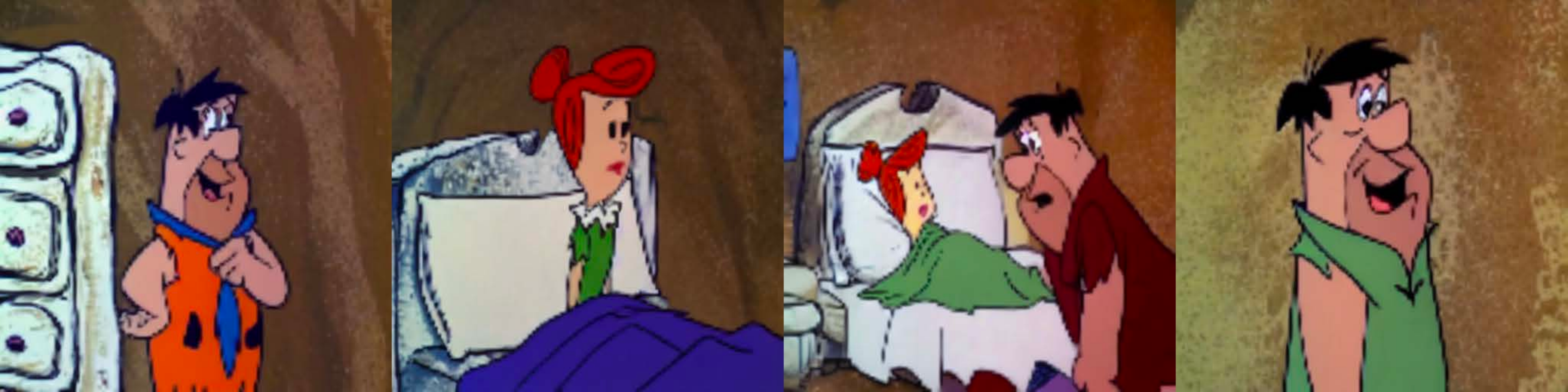}
\caption{Visual story synthesized by AR-LDM.}
\label{fig:dicussion_arldm}
\end{subfigure}
\caption{A failure case of AR-LDM.}
\label{fig:dicussion}
\end{figure}

\setlength\parindent{0pt}
\footnotesize{\textit{
1. Mr Slate who is really Fred is in his bedroom talking with his hand on his chest and then he points his finger.\\
2. Wilma is in the bedroom. She is sitting in the bed.\\
3. Fred and Wilma are in the bedroom. Wilma is beneath a blanket while Fred is beside the bed, speaking to her.\\
4. Fred is in the living room. He is talking and wearing a green shirt.
}}
\setlength\parindent{1pc}

\newpage
\section{Ethical Statement}
For the two datasets introduced in previous works, PororoSV and FlintstonesSV, images are unable to be confused with real images due to the fact that they are cartoons. As for the newly introduced VIST dataset, we follow a CreativeML Open RAIL-M license (\url{https://huggingface.co/spaces/CompVis/stable-diffusion-license}) used by Stable Diffusion~\cite{ldm}. Also, based on the pre-trained Stable Diffusion weight, we share the same biases and content acknowledgment as follows:

\paragraph{Biases and Content Acknowledgment}
Despite how impressive being able to turn text into image is, beware to the fact that this model may output content that reinforces or exacerbates societal biases, as well as realistic faces, pornography and violence. The model was trained on the LAION-5B dataset, which scraped non-curated image-text-pairs from the internet (the exception being the removal of illegal content) and is meant for research purposes.

\newpage
\section{Annotation Instructions for Human Evaluation}
\label{sec:annotation_instructions_for_human_evaluation}
\setlength\parindent{0pt}
The annotation instructions for the human evaluation are provided here:\\

\texttt{We have collected synthesized visual stories from different generative models. We would like you to help us choose the preferred ones from these stories regarding three orthogonal criteria, visual quality, relevance, and consistency. Visual stories from (anonymous) generative models have been shuffled on EACH ROW, so you the annotator cannot know which model they come from.\\\\
PLEASE READ THESE INSTRUCTIONS IN FULL.\\\\
Annotation Rules for Visual Quality:
\begin{itemize}
    \item You are given two single images (without caption) synthesized by two different generative models to describe one specific caption. Choose the synthesized image only according to which one is in higher visual quality.
    \item What is the best answer? Make a decision based on (a) the scenes, entities, and characters are logical and not obviously unreasonable, like misplacement or clipping; (b) the image is clear, rich in detail, and realistic in color.
    \item If two images provide the same visual quality by your judgment, and there is no clear winner, you may rank them the tie, but please only use this sparingly.
\end{itemize}
Annotation Rules for Relevance:
\begin{itemize}
    \item You are given one specific caption and two images synthesized by two different generative models to describe the given caption. Choose the synthesized image only according to which one is more relevant to the given caption.
    \item What is the best answer? Make a decision based on (a) the image conforms well to the content of the caption; (b) the image reflects the details and objects described in the caption instead of omitting them. Overall, use your best judgment to choose answers based on being the most relevant image, which we define as one which is at least somewhat correct, and minimally informative about what the caption is describing.
    \item Images in higher quality are not always the best. An image that is more relevant to the given caption may be better than one in higher quality, if they are at least as correct and informative.
    \item If two images provide the same relevance by your judgment, and there is no clear winner, you may rank them the tie, but please only use this sparingly.
\end{itemize}
Annotation Rules for Consistency:
\begin{itemize}
    \item You are given two visual stories synthesized by two different generative models to describe specific captions. Choose the synthesized visual story only according to which one is more consistent.
    \item What is the best answer? Make a decision based on the scenes as well as recurring entities and characters are consistent across frames.
    \item Visual stories in higher quality are not always the best. A visual story that is more consistent across frames may be better than one in higher quality, if they are at least as correct and informative.
    \item If two visual stories provide the same consistency by your judgment, and there is no clear winner, you may rank them the tie, but please only use this sparingly.
\end{itemize}
}
\end{document}